\documentclass{article}

\usepackage{arxiv}

\usepackage[utf8]{inputenc} % allow utf-8 input
\usepackage[T1]{fontenc}    % use 8-bit T1 fonts
\usepackage{hyperref}       % hyperlinks
\usepackage{url}            % simple URL typesetting
\usepackage{booktabs}       % professional-quality tables
\usepackage{amsfonts}       % blackboard math symbols
\usepackage{nicefrac}       % compact symbols for 1/2, etc.
\usepackage{microtype}      % microtypography
\usepackage{lipsum}
\usepackage{graphicx}

\usepackage{booktabs}       % 用于 \toprule, \midrule, \bottomrule 等命令
\usepackage{tabularx}       % 用于 tabularx 环境和 X 列类型
\usepackage{multirow}       % 用于 \multirow 命令
\usepackage{makecell}       % 用于 \makecell 命令
\usepackage{array}          % 用于 >{\centering\arraybackslash}X 列格式
\usepackage{multicol}
\usepackage{siunitx}  % 提供 S 列类型，用于自动对齐数字
\usepackage{enumitem}
\usepackage{bm, booktabs, amsmath}
\usepackage{hyperref}
\usepackage[capitalize]{cleveref}
\crefname{section}{Sec.}{Secs.}
\Crefname{section}{Section}{Sections}
\Crefname{table}{Table}{Tables}
\crefname{table}{Tab.}{Tabs.}

\title{LocusGS: Spatially Grounded Tokens for Feed-Forward 3D Gaussian Splatting}
\author{
  Wenyu Li, Sidun Liu, Tongrui Hu, Peng Qiao and Yong Dou \\
  National University of Defence Technology \\
  Changsha, China \\
  \texttt{\{wenyu18, liusidun, tongruihu, pengqiao, yongdou\}@nudt.edu.cn} \\
}

\begin{document}

\maketitle

\begin{abstract}
Recent query-based feed-forward 3DGS methods represent a scene using
learnable queries, each aggregating multi-view evidence and decoding a
group of Gaussians.
Ideally, different queries should specialize in coherent local regions
of the scene.
However, we observe that Gaussians decoded from the same query often
scatter across distant scene regions, resulting in weak query-level
spatial coherence and poor alignment with the scene structure.
We attribute this behavior to the purely latent representation of
existing Gaussian queries.
To address this limitation, we introduce LocusGS, which augments each Gaussian query with a 3D
anchor state consisting of a center and a support radius.
The anchor state is progressively refined across decoder layers and is
used throughout query interaction, multi-view feature aggregation, and
Gaussian generation.
Specifically, an anchor-to-ray geometric bias guides each query toward
spatially relevant image observations, while anchor-centered decoding
organizes its Gaussians within a local region.
Experiments on novel view synthesis benchmarks show that
LocusGS improves rendering quality over query-based Gaussian token
baselines under the same Gaussian budget. 
Further analysis shows that the learned anchors form
coherent spatial layouts and lead to more structured Gaussian
distributions, demonstrating that explicit anchor states improve the spatial
organization.
Our project page: \url{https://leo-frank.github.io/LocusGS_viewer}
\end{abstract}

\section{Introduction}

\begin{figure}[t]
    \centering
    \includegraphics[width=0.7\textwidth]{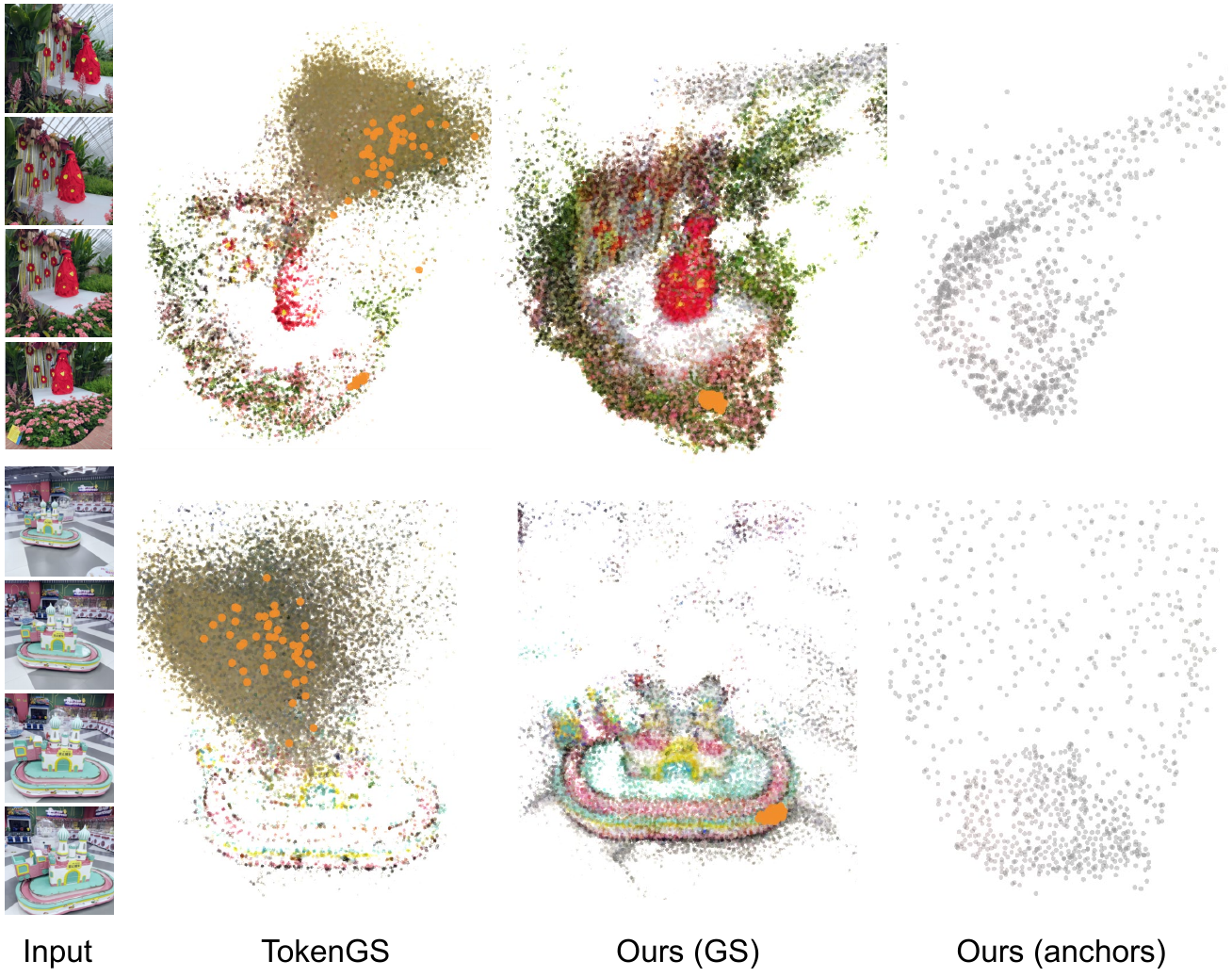}
    \caption{
        \textbf{Spatial grounding improves both global Gaussian organization and token-level locality.}
        The orange points highlight all 64 Gaussians decoded from a single representative token.
        Under the same input and Gaussian budget, TokenGS produces diffuse scene structures, with the Gaussians from one token scattered across a broad spatial region.
        LocusGS instead reconstructs Gaussians that better follow the scene geometry, and its learned anchors form a coarse spatial scaffold for organizing local Gaussian generation.
        }
    \label{fig:global_gs_compare}
\end{figure}

Reconstructing 3D scenes~\cite{schonberger16structure-from-motion,schonberger16pixelwise,mildenhall2021nerf} from images is a fundamental problem in computer vision, with broad applications in robotics, augmented reality, and embodied perception. 
3D Gaussian Splatting~\cite{kerbl3Dgaussians} has shown that a scene can be represented by a set of Gaussian primitives and rendered with high fidelity and efficiency.
However, the original Gaussian Splatting pipeline typically requires costly iterative fitting for each new scene. 
To reduce this cost, recent feed-forward methods attempt to predict 3D Gaussian representations directly from input views, enabling faster and more scalable 3D scene reconstruction.

\begin{figure}[t]
    \centering
    \includegraphics[width=\columnwidth]{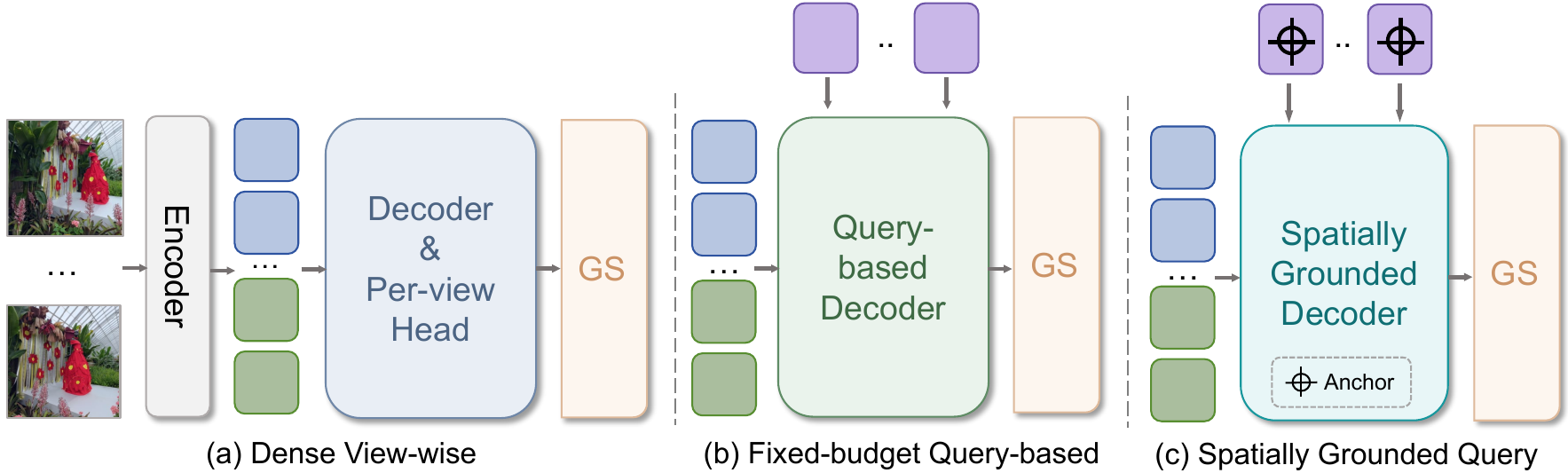}
    \caption{
        \textbf{Conceptual comparison of three feed-forward 3DGS paradigms.}
        (a) Dense view-wise prediction decodes multi-view image features and uses a per-view head to produce view-dependent Gaussian splats.
        (b) Fixed-budget query-based generation introduces learnable queries
            to decode a set of Gaussians whose size is independent of image
            resolution and input view count.
        (c) Our spatially grounded query paradigm further equips queries with explicit spatial anchors, enabling anchor-aware decoding and more structured Gaussian generation.
        }
    \label{fig:paradigm_comparison}
\end{figure}

Among feed-forward 3DGS methods, query-based prediction has emerged as an alternative to dense, pixel-aligned Gaussian regression.
Unlike dense methods that predict Gaussian primitives directly from image-grid features~\cite{charatan23pixelsplat:,chen24mvsplat:,xu2024depthsplat}, query-based methods represent the scene using a fixed set of learnable queries~\cite{tokengs2026,globalsplat}.
Each query aggregates multi-view image evidence through a Transformer decoder and is subsequently decoded into a group of 3D Gaussians.
This fixed query set decouples the Gaussian budget from both image resolution and the number of input views, as illustrated in \Cref{fig:paradigm_comparison}.
However, removing the dense image-to-Gaussian correspondence also leaves the spatial role of each query implicit.

Ideally, query tokens should exhibit clear spatial specialization.
Each token should generate a compact Gaussian group within a coherent local region, while different tokens cover complementary parts of the scene.
However, as shown in \Cref{fig:global_gs_compare}, Gaussians decoded from the same token often scatter across distant scene regions, resulting in a spatially diffuse distribution that poorly aligns with the scene structure.
These observations suggest that existing latent-only query tokens lack a clear notion of spatial responsibility.
We argue that this failure stems from the lack of an explicit spatial state: a latent query does not specify where it operates in 3D or how large a region it covers.
Consequently, neither the multi-view evidence aggregated by a query nor
the Gaussians decoded from it is explicitly constrained to a coherent
local region.

To establish such spatial specialization, we propose LocusGS, which augments each Gaussian token with an explicit 3D anchor state consisting of a center and a support radius.
The center specifies the token's current 3D location, while the radius defines the extent of its local support.
The anchor state is used throughout the decoding phase: It guides token-to-image cross-attention, provides spatial cues for interactions among Gaussian tokens, and serves as a local reference for Gaussian generation.
The anchor center and radius are progressively refined across decoder layers, enabling each token to adapt its spatial location and support to the input scene.
In this way, LocusGS spatially grounds each query and encourages it to specialize in a compact local region.

Experiments on novel view synthesis benchmarks show that LocusGS
improves rendering quality over query-based Gaussian token baselines
under identical token and Gaussian budgets. More importantly, the proposed anchor
states lead to a more structured 3D organization of the predicted
Gaussians. Our analysis shows that LocusGS produces fewer scattered
token-associated primitives, encourages the Gaussians decoded from the
same token to form spatially compact local groups, and learns anchors that form a
coherent spatial scaffold over the reconstructed scene. These results
indicate that explicit 3D anchor states provide not only better
reconstruction accuracy, but also a more spatially meaningful query
representation for feed-forward Gaussian reconstruction.

\section{Related Work}
\subsection{Feed-forward 3D Gaussian Splatting}
Feed-forward 3D Gaussian Splatting predicts scene representations
directly from sparse input images, avoiding costly per-scene
optimization.
PixelSplat~\cite{charatan23pixelsplat:} predicts Gaussians through
epipolar feature aggregation and depth estimation, while
MVSplat~\cite{chen24mvsplat:} introduces cost-volume-based multi-view
reasoning.
DepthSplat~\cite{xu2024depthsplat} further incorporates monocular depth
priors, and subsequent works extend feed-forward reconstruction to
pose-free settings~\cite{kang2025selfsplat,ye2024noposplat}.
These methods largely generate Gaussians from dense image-aligned
features, coupling the primitive budget with image resolution and input
view count.

\subsection{Grid-Decoupled and Query-based Gaussian Reconstruction}
To reduce the redundancy of dense Gaussian prediction, recent methods
adopt selective primitive generation~\cite{wang2024freesplat,lin2025longsplat},
feature or token compression~\cite{wang2025zpressor,ziwen2025llrm},
or cross-view token fusion~\cite{tokensplat}.
Other methods leverage geometric priors from pretrained 3D
reconstruction models to merge redundant pixel-aligned
predictions~\cite{jiang2025anysplat} or construct sparse 3D
anchors~\cite{zhang2026anchorsplat}.
These methods reduce redundancy while retaining an image-aligned or
externally initialized prediction structure.
Query-based methods instead decouple the scene representation itself
from the input grid.
TokenGS~\cite{tokengs2026} introduces learnable Gaussian queries, each
of which aggregates multi-view image features and decodes a group of
Gaussian primitives, making the output budget independent of image resolution
and view count.
Several contemporaneous methods share this high-level query-based
formulation, while differing in their input assumptions, feature
backbones, attention architectures, and Gaussian decoding
strategies~\cite{globalsplat,an2025c3g}.
GlobalSplat\cite{globalsplat} adopts disentangled geometry and appearance
branches, whereas C3G\cite{an2025c3g} uses a pretrained VGGT encoder and decodes one
Gaussian per query.
Nevertheless, these methods primarily represent each query as a latent
feature.
LocusGS instead equips each query with an explicit 3D state that is progressively refined and directly involved in both feature aggregation and Gaussian decoding. This design is conceptually related to DAB-DETR~\cite{liu2022dabdetr}, which augments object queries with dynamically refined anchor boxes. While DAB-DETR uses such anchors for 2D object localization, LocusGS maintains a center-and-radius state in 3D space to spatially ground scene queries throughout reconstruction.
LocusGS is also related to Scaffold-GS~\cite{scaffoldgs}, which organizes local Gaussians around sparse 3D anchors, while our anchors serve as query states and are progressively refined from multi-view image features in a feed-forward framework.

\section{Method}

\subsection{Preliminaries: Query-based Feed-forward Gaussian Splatting}

Recent query-based feed-forward Gaussian Splatting methods formulate
3DGS prediction as a set-to-set decoding problem. Given multi-view images
\(\mathcal{I}=\{I_v\}_{v=1}^{V}\) with corresponding camera poses
\(\mathcal{P}\), an image encoder first extracts multi-view visual
features \(\mathbf{F}\), which serve as the memory for the decoder.
The decoder maintains a set of \(N\) learnable Gaussian queries,
initialized as
\[
\mathbf{Q}^{0} =
[\mathbf{q}_1^{0}, \mathbf{q}_2^{0}, \ldots, \mathbf{q}_N^{0}]^\top
\in \mathbb{R}^{N\times d},
\]
where \(\mathbf{q}_i^{0}\in\mathbb{R}^{d}\) denotes the initial feature
of the \(i\)-th query. These queries are progressively updated
by Transformer decoder blocks:
\[
\mathbf{Q}^{l+1}=D(\mathbf{Q}^{l},\mathbf{F}).
\]
Each decoder block typically alternates \textit{self-attention} among Gaussian
queries and \textit{cross-attention} to the encoded multi-view image features,
allowing the queries to aggregate scene evidence from the input views.
After \(L\) decoder layers, each final query \(\mathbf{q}_i^{L}\) is
mapped by a Gaussian prediction head to a local group of $K$ Gaussian
primitives:
\[
\mathcal{G}_i =
\{(\mathbf{x}_{i,k}, \mathbf{s}_{i,k}, \mathbf{R}_{i,k},
\mathbf{c}^{G}_{i,k}, \alpha_{i,k})\}_{k=1}^{K},
\]
where \(\mathbf{x}\), \(\mathbf{s}\), \(\mathbf{R}\), \(\mathbf{c}^{G}\),
and \(\alpha\) denote the Gaussian center, scale, rotation, color, and
opacity, respectively. 
The Gaussian groups decoded from all tokens are combined to form the
scene representation
\(\mathcal{G}=\bigcup_{i=1}^{N}\mathcal{G}_i\).

\subsubsection{Limitations of Existing Query-based Methods}
Despite the efficiency of fixed-budget prediction, existing
query-based methods exhibit limited \emph{spatial specialization}.
Ideally, each query should focus on a coherent local region, aggregate
the multi-view evidence associated with that region, and decode a
spatially compact group of Gaussians, while different queries cover
complementary parts of the scene.
However, as illustrated in \Cref{fig:global_gs_compare}, 
the Gaussians decoded from an individual query
are often spatially diffuse and may spread across distant scene
regions.
As a result, the token-wise Gaussian groups show weak spatial
coherence, and their organization does not closely follow the scene
structure.
% We attribute this behavior to the lack of an \emph{explicit 3D spatial
% state}: a latent query does not specify where it operates in 3D or how
% large a region it covers.
% Consequently, the multi-view evidence aggregated by a query and the
% Gaussians decoded from it are not tied to the same coherent local
% region.
% This limitation motivates us to augment each query with an explicit 3D
% anchor state, enabling geometry-aware feature aggregation and locally
% organized Gaussian generation.

\begin{figure}[t]
    \centering
    \includegraphics[width=0.7\columnwidth]{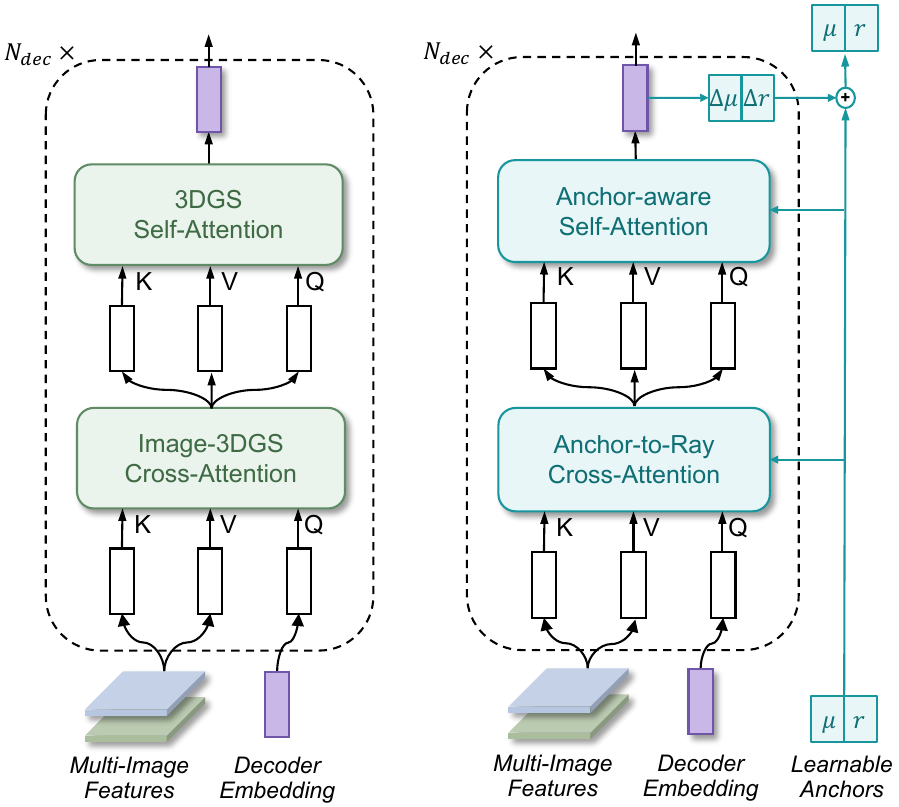}
    \caption{
        \textbf{From implicit Gaussian tokens to explicit anchor tokens.}
        Left: standard query-based methods uses learnable 3DGS tokens as implicit embeddings, where self-attention and image-to-3DGS cross-attention are mainly driven by token and image features.
        Right: LocusGS augments each decoder token with a learnable 3D anchor state $(\mu,r)$.
        The anchor state provides spatial cues for token self-attention and guides cross-view feature aggregation through anchor-aware cross-attention.
        Across decoder layers, the anchor state is progressively refined and used for anchor-centered Gaussian prediction, turning Gaussian tokens into spatially grounded reconstruction queries.
        }
    \label{fig:method}
\end{figure}

\subsection{LocusGS}

\subsubsection{Our Insight}
We argue that the weak spatial coherence observed above stems from \emph{how Gaussian
queries are represented}.
Existing methods typically represent each query
solely as a latent feature.
Such a representation does not explicitly specify \emph{where the query
operates in 3D space} or \emph{how large a region it should cover}.
Consequently, the multi-view evidence aggregated by a query and the
Gaussians decoded from it are not associated with the coherent
local region.
This observation motivates us to equip each query with an explicit 3D
anchor state, providing a spatial reference for feature aggregation 
and locally organized Gaussian prediction.

\subsubsection{Overview}
Given posed multi-view images, LocusGS predicts a fixed-budget set of
3D Gaussian primitives in a single forward pass. The input images are encoded
into multi-view image tokens, with camera parameters represented by
patch-level Plücker rays.
LocusGS augments each query with an explicit 3D anchor state that
specifies its current 3D reference location and support radius. The
decoder uses these anchors to guide cross-view feature aggregation,
progressively refines the anchor states, and decodes Gaussian primitives
as local offsets from the refined anchors.

\subsubsection{Explicit 3D Anchor States for Queries}
In existing query-based Gaussian reconstruction, each query is represented as a learnable high-dimensional embedding. Such an embedding acts as a content query in the Transformer decoder.
Different from standard query-based methods, LocusGS augments every query
with an explicit 3D anchor state. For the \(i\)-th query at decoder layer
\(l\), we denote its token feature as
\(\mathbf{q}_i^l \in \mathbb{R}^{d}\) and its anchor state as
\[
\mathbf{a}_i^l
=
(\boldsymbol{\mu}_i^l,r_i^l)
\]
where \(\boldsymbol{\mu}_i^l\in\mathbb{R}^{3}\) is the center and
\(r_i^l\in\mathbb{R}_{+}\) is its spatial support radius. The token
feature encodes appearance and reconstruction cues, while the anchor
state provides an explicit spatial descriptor: the center specifies the
current 3D reference location of the query, and the radius defines its
local support region.

To ensure a positive support radius, in practice, we maintain an unconstrained
radius parameter \(\rho_i^l\in\mathbb{R}\) and obtain the actual radius as
\begin{equation}
r_i^l
=
\operatorname{softplus}(\rho_i^l)
+
\epsilon
\label{eq:radius_parameterization}
\end{equation}
where \(\epsilon>0\) is a small constant for numerical stability.
The initial anchor parameters
\(\{\boldsymbol{\mu}_i^0,\rho_i^0\}_{i=1}^{N}\) are learnable and shared
across scenes, with the anchor centers randomly initialized in the
normalized 3D scene space. 
% During decoding, the anchor state provides spatial cues, adapts the geometric attention range, and
% defines a local coordinate frame for Gaussian prediction.

\subsubsection{Anchor-Guided Decoder}
As in conventional query-based Gaussian reconstruction methods, the decoder updates query tokens through alternating self-attention and cross-attention. 
Here, LocusGS makes each decoder layer aware of the current anchor states.

For \textit{self-attention}, we derive an anchor positional embedding
\(\mathbf{p}_i^l = \mathrm{MLP}(\mathrm{PE}(\boldsymbol{\mu}_i^l))\)
from the current anchor center, where \(\mathrm{PE}(\cdot)\) denotes
sinusoidal positional encoding. We inject this embedding into the token
feature before self-attention:
\begin{equation}
\tilde{\mathbf{q}}_i^{\,l}
=
\mathbf{q}_i^{\,l}
+
\mathbf{p}_i^{\,l}.
\end{equation}
This spatial conditioning allows interactions among Gaussian queries to
depend on their current 3D anchor locations, rather than only on their
latent token features.

For \textit{cross-attention}, we aim to make each query preferentially
aggregate multi-view evidence consistent with its current 3D support.
To assess this geometric consistency, we measure the distance between
the query anchor \(\boldsymbol{\mu}_i^{\,l}\) and the camera ray
\(\boldsymbol{\ell}_j\) associated with each image token.
Image tokens whose rays pass closer to the anchor are considered more
relevant, while the support radius \(r_i^{\,l}\) controls the spatial
extent of this preference.
We incorporate this relation as an anchor-to-ray geometric bias in
standard content-based cross-attention.
Concretely, for the camera ray \(\boldsymbol{\ell}_j\) associated with
the \(j\)-th image token, we define the anchor-to-ray geometric bias as
\begin{equation}
b_{ij}^{\,l}
=
-
\tfrac{1}{2}
\left(
\frac{
D\left(\boldsymbol{\mu}_i^{\,l}, \boldsymbol{\ell}_j\right)
}{
\sigma_0 r_i^{\,l}
}
\right)^2 ,
\end{equation}
where \(D\left(\boldsymbol{\mu}_i^{\,l}, \boldsymbol{\ell}_j\right)\) denotes the shortest
Euclidean distance from the 3D point \(\boldsymbol{\mu}_i^{\,l}\) to the camera
ray \(\boldsymbol{\ell}_j\), \(\sigma_0\) is a
fixed bandwidth hyperparameter, and \(r_i^{\,l}\) is the anchor support radius. This
bias assigns higher scores to image tokens whose rays are closer to the
anchor, while the radius controls how broad the geometric support is.

The geometric bias is added to the standard content-based attention
logits. Let \(\bar{\mathbf{q}}_i^{\,l}\) be the attention query projected
from the Gaussian query feature \(\mathbf{q}_i^{\,l}\). The cross-attention
weights are computed as
\begin{equation}
\alpha_{ij}^{\,l}
=
\operatorname{softmax}_{j}
\left(
\frac{
\left(\bar{\mathbf{q}}_i^{\,l}\right)^{\top}\mathbf{k}_j
}{
\sqrt{d}
}
+
\gamma b_{ij}^{\,l}
\right),
\end{equation}
where \(\mathbf{k}_j\) is the projected key of the \(j\)-th image token
and \(d\) is the feature dimension per attention head.
Here, \(\gamma=\operatorname{softplus}(\tilde{\gamma})\geq 0\) is a
learnable scale, ensuring that the geometric bias consistently penalizes
rays farther from the anchor.
The updated
query feature is obtained by aggregating image values:
\begin{equation}
\hat{\mathbf{q}}_i^{\,l}
=
\sum_j
\alpha_{ij}^{\,l}
\mathbf{v}_j .
\end{equation}
In this way, each query aggregates multi-view evidence by jointly
considering content similarity and anchor-to-ray geometric consistency.

\begin{figure*}[t]
    \centering
    \includegraphics[width=\textwidth]{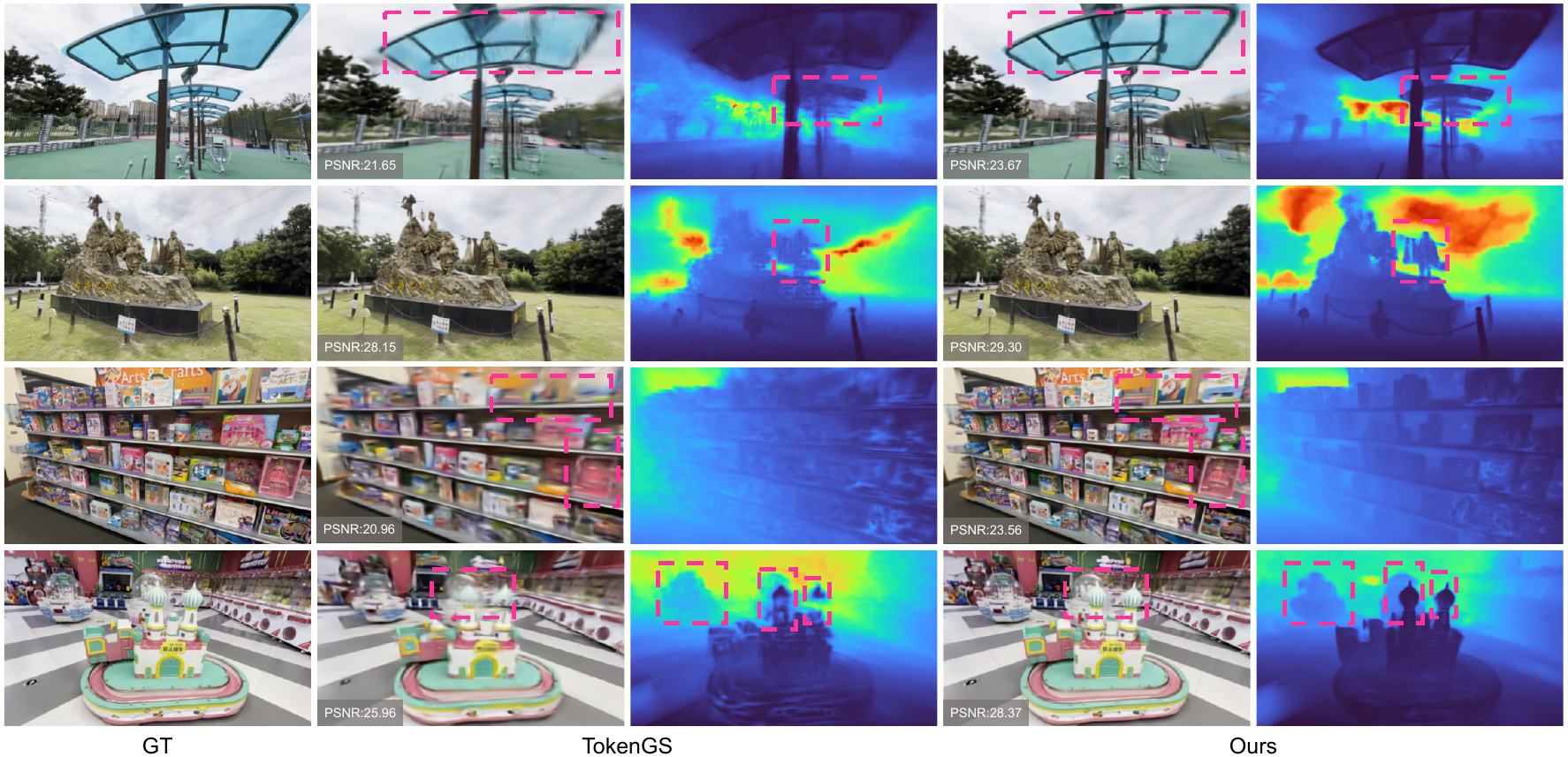}
    \caption{
        \textbf{Qualitative comparison on DL3DV novel-view synthesis.}
        LocusGS produces sharper renderings and more coherent depth structures, especially in texture-rich regions and cluttered scenes.
        }
    \label{fig:nvs_and_depth}
\end{figure*}

\subsubsection{Dynamic Anchor Refinement}
After each decoder layer, the updated token feature predicts residual
updates for the anchor center and the unconstrained radius parameter:
\begin{equation}
\Delta \boldsymbol{\mu}_i^{\,l}
=
f_{\mu}\left(\mathbf{q}_i^{\,l+1}\right),
\qquad
\Delta \rho_i^{\,l}
=
f_{\rho}\left(\mathbf{q}_i^{\,l+1}\right),
\end{equation}
where \(f_{\mu}(\cdot)\) and \(f_{\rho}(\cdot)\) are lightweight
prediction heads. The anchor parameters are refined as
\begin{equation}
\boldsymbol{\mu}_i^{\,l+1}
=
\boldsymbol{\mu}_i^{\,l}
+
\Delta \boldsymbol{\mu}_i^{\,l},
\qquad
\rho_i^{\,l+1}
=
\rho_i^{\,l}
+
\Delta \rho_i^{\,l}.
\end{equation}
The refined support radius \(r_i^{\,l+1}\) is then obtained from
\(\rho_i^{\,l+1}\) using the parameterization in
\Cref{eq:radius_parameterization}.
This layer-wise refinement enables each query to adapt its 3D location
and support radius according to the multi-view input images.

\subsubsection{Anchor-Centered Gaussian Decoding}
After the final decoder layer, each token is decoded into a small group of Gaussian primitives. Instead of predicting Gaussian positions in a fully unconstrained manner, LocusGS predicts local offsets relative to the final anchor center. For the $k$-th Gaussian generated by the $i$-th token, the decoder first predicts a local offset $\boldsymbol{\delta}_{i,k}$ from the final token feature:
\begin{equation}
\{\boldsymbol{\delta}_{i,k}\}_{k=1}^{K}
=
f_{\delta}(\boldsymbol{q}_i^L)
\end{equation}
where \(L\) is the number of decoder layers and
\(f_{\delta}(\cdot)\) denotes the position branch of the Gaussian
prediction head. 
The final Gaussian center is then obtained by anchoring this local offset around the refined anchor center:
\begin{equation}
\boldsymbol{\mu}_{i,k}^{G}
=
\boldsymbol{\mu}_i^{\,L}
+
r_i^{\,L} \boldsymbol{\delta}_{i,k}
\end{equation}
Here, $\boldsymbol{\mu}_i^{\,L}$ provides the final 3D reference location of the token, while $r_i^{\,L}$ controls the spatial range of its generated Gaussians. This formulation encourages the Gaussians decoded from the same token to form a local 3D group around the anchor state, rather than scattering freely in the scene.

The remaining Gaussian attributes, including scale, rotation, color, and
opacity, are predicted from the final token feature using the standard
token-based prediction head as in TokenGS~\cite{tokengs2026}. The final
Gaussian primitives are formed by combining these attributes with the
anchor-centered Gaussian centers.

\subsubsection{Multi-layer Rendering Supervision}
Since LocusGS progressively refines the anchor states, we supervise intermediate decoder layers in addition to the final output.
Let $\mathcal{S}=\{l_1,\ldots,l_M\}$ denote the supervised layers, with the final layer always included.
At each layer $l_m$, the current Gaussian tokens and anchor states are decoded into an intermediate Gaussian set $\mathcal{G}^{\,l_m}$ and rendered as
$\hat{\mathbf{I}}_t^{\,l_m}
=
\mathcal{R}(\mathcal{G}^{\,l_m},\Pi_t)$.
Following TokenGS, we combine the image reconstruction loss with visibility regularization on both the decoded Gaussian centers and anchor centers:
% \begin{equation}
% \mathcal{L}^{\,l_m}
% =
% \mathcal{L}_{\mathrm{rec}}^{\,l_m}
% +
% \lambda_{\mathrm{vis}}\mathcal{L}_{\mathrm{vis}}^{G,l_m}
% +
% \lambda_{\mathrm{anchor}}\mathcal{L}_{\mathrm{vis}}^{\mathrm{Anchor},l_m}.
% \end{equation}
\begin{equation}
\mathcal{L}^{l_m}
=
\mathcal{L}_{\mathrm{rec}}^{l_m}
+
\lambda_G
\mathcal{L}_{\mathrm{vis}}~\!\bigl(
\{\boldsymbol{\mu}_{i,k}^{G,l_m}\}
\bigr)
+
\lambda_A
\mathcal{L}_{\mathrm{vis}}~\!\bigl(
\{\boldsymbol{\mu}_{i}^{l_m}\}
\bigr)
\end{equation}
The overall training objective is
\begin{equation}
\mathcal{L}
=
\sum_{m=1}^{M}
w_m
\mathcal{L}^{\,l_m},
\qquad
w_m
=
\frac{m}{\sum_{n=1}^{M}n}
\end{equation}
This supervision directly regularizes intermediate anchor refinement while assigning larger weights to later decoding stages.
Detailed loss definitions are provided in the supplementary material.

\section{Experiments}

\subsection{Experimental Setup}

We evaluate LocusGS on two large-scale scene-level datasets, RE10K~\cite{re10k} and DL3DV~\cite{ling2024dl3dv}, following the reconstruction settings commonly used in prior feed-forward Gaussian methods~\cite{tokengs2026}. For RE10K, we report two-view reconstruction results at $256 \times 256$ resolution. For DL3DV, we train the base model at $256 \times 256$ and further finetune it at $448 \times 256$ for higher-resolution evaluation. The DL3DV model is trained with four views and tested with varying numbers of context views to evaluate cross-view generalization. 
% To reduce memory cost in the high-resolution multi-view setting, we use a compact backbone with 3 encoder layers and 12 decoder layers. 
We use TokenGS as the primary query-based baseline 
because it provides a clean controlled setting: it uses posed inputs
and learnable Gaussian queries without an additional pretrained
geometric reconstruction backbone.
Comparing under matched token and Gaussian
budgets allows us to isolate the effect of explicit spatial grounding;
a detailed discussion of other concurrent token-based methods is
provided in supplementary material.

\subsection{Main Results}

\Cref{tab:re10k_two_view,tab:dl3dv_view_generalization} compare our method with representative feed-forward 3DGS baselines. 
On RealEstate10K~\cite{re10k}, our method consistently outperforms TokenGS under the same token and Gaussian budgets. 
Both the 1024-token and 4096-token variants achieve higher PSNR and SSIM with lower LPIPS, showing that our design improves reconstruction quality without increasing the number of predicted Gaussians. 
Notably, the 1024-token variant already surpasses GS-LRM~\cite{gslrm2024} in PSNR and SSIM while using only half the number of Gaussians, and the 4096-token variant achieves the best PSNR and SSIM among all compared methods.

On DL3DV~\cite{ling2024dl3dv}, our model is trained with 4 input views and directly evaluated under 2-, 4-, and 6-view settings. 
Compared with TokenGS, our method consistently improves reconstruction quality across all view settings using the same number of Gaussians. 
This suggests that the proposed design not only improves the reconstruction quality under the training configuration, but also generalizes well to unseen context lengths. 
Moreover, unlike pixel-aligned methods such as MVSplat and DepthSplat,
whose number of Gaussians increases with the number of input views,
our method maintains a fixed, view-count-independent Gaussian budget.
\Cref{fig:nvs_and_depth} provides qualitative comparisons, where our method shows clearer renderings and more consistent geometric structures compared with TokenGS.

\begin{table}[t]
\centering
\caption{
Evaluations on the DL3DV~\cite{ling2024dl3dv} dataset with different numbers of input views.
Our model is trained with 4 input views and directly evaluated under 2-, 4-, and 6-view settings.
Resolution is $448 \times 256$.
}
\label{tab:dl3dv_view_generalization}
\footnotesize
\begin{tabular}{lccccc}
\toprule
Method & \#Views & PSNR$\uparrow$ & SSIM$\uparrow$ & LPIPS$\downarrow$ & \#GS \\
\midrule

MVSplat
& \multirow{4}{*}{2}
& 17.54 & 0.529 & 0.402 & 229K \\
DepthSplat
&
& 19.31 & 0.615 & \textbf{0.310} & 229K \\
TokenGS (4096 tok)
&
& 19.58 & 0.615 & 0.429 & 262K \\
\textbf{Ours} (4096 tok)
&
& \textbf{20.90} & \textbf{0.678} & 0.377 & 262K \\

\midrule

MVSplat
& \multirow{4}{*}{4}
& 21.63 & 0.721 & 0.233 & 458K \\
DepthSplat
&
& 23.12 & 0.780 & \textbf{0.178} & 458K \\
TokenGS (4096 tok)
&
& 23.44 & 0.757 & 0.312 & 262K \\
\textbf{Ours} (4096 tok)
&
& \textbf{24.80} & \textbf{0.812} & 0.248 & 262K \\

\midrule

MVSplat
& \multirow{4}{*}{6}
& 22.93 & 0.775 & 0.193 & 688K \\
DepthSplat
&
& 24.19 & 0.823 & \textbf{0.147} & 688K \\
TokenGS (4096 tok)
&
& 24.16 & 0.770 & 0.296 & 262K \\
\textbf{Ours} (4096 tok)
&
& \textbf{25.78} & \textbf{0.836} & 0.225 & 262K \\

\bottomrule
\end{tabular}
\end{table}

\begin{table}[t]
\centering
\caption{
Reconstruction performance with two input views on RealEstate10K~\cite{re10k}.
Resolution is $256 \times 256$.
}
\label{tab:re10k_two_view}
\footnotesize
\begin{tabular}{lcccc}
\toprule
Method & PSNR$\uparrow$ & SSIM$\uparrow$ & LPIPS$\downarrow$ & \#GS \\
\midrule
MVSplat
& 26.39 & 0.869 & 0.128 & 131K \\
DepthSplat
& 27.47 & 0.889 & \textbf{0.114} & 131K \\
GS-LRM
& 28.10 & 0.892 & \textbf{0.114} & 131K \\

\midrule
TokenGS (1024 tok)
& 28.02 & 0.896 & 0.147 & 66K \\
\textbf{Ours} (1024 tok)
& 28.50 & 0.909 & 0.135 & 66K \\

\midrule
TokenGS (4096 tok)
& 28.41 & 0.903 & 0.135 & 262K \\
\textbf{Ours} (4096 tok)
& \textbf{28.89} & \textbf{0.916} & 0.124 & 262K \\

\bottomrule
\end{tabular}
\end{table}

\begin{figure}[t]
    \centering
    \includegraphics[width=0.8\columnwidth]{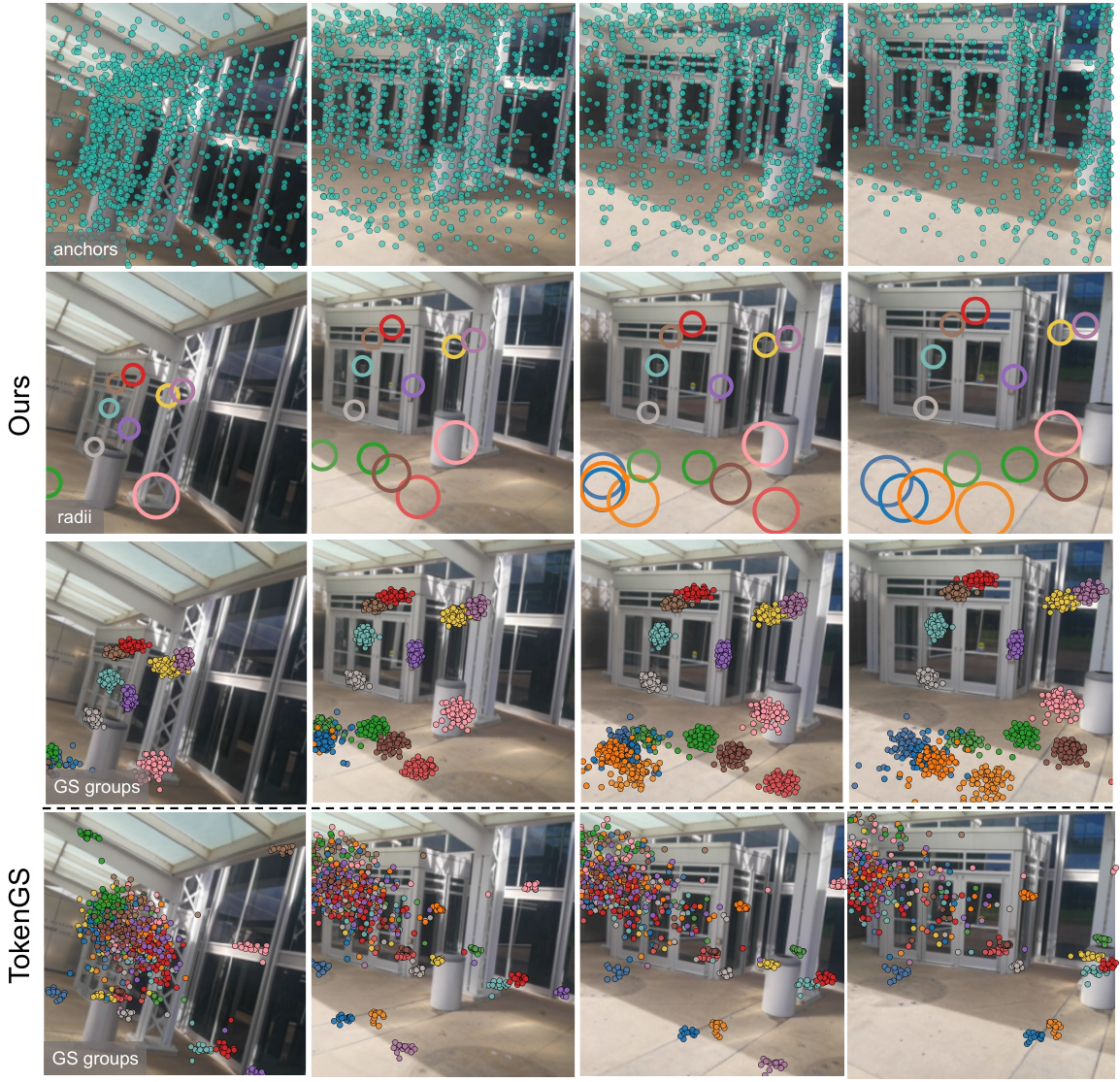}
    \caption{
        \textbf{Visualization of anchor supports and token-level Gaussian compactness.}
        The first three rows show LocusGS: all projected anchors, selected projected radii,
        and Gaussian groups decoded from the selected anchor tokens.
        The last row shows Gaussian groups decoded from TokenGS tokens.
        LocusGS learns adaptive spatial supports and encourages each token to decode a more compact local Gaussian group.
        }
    \label{fig:token_compactness}
\end{figure}
% \begin{figure}[t]
%     \centering
%     \includegraphics[width=\columnwidth]{Figures/3D_compactness.png}
%     \caption{
%         \textbf{3D visualization of token-level compactness.}
%         The Gaussians decoded from a selected token are highlighted in orange.
%         Our method produces a more compact and spatially coherent Gaussian group.        }
%     \label{fig:token_compactness_3d}
% \end{figure}

\begin{table}[t]
\centering
\caption{
Quantitative token-level Gaussian dispersion on the test sets.
We report the statistics of \(C_{\mathrm{centroid}}\); lower is better.
}
\label{tab:quan_token_compactness}
\footnotesize
\begin{tabular}{@{}llccc@{}}
\toprule
Dataset & Method
& Mean$\downarrow$
& Variance$\downarrow$
& Median$\downarrow$ \\
\midrule

\multirow{2}{*}{RE10K}
& TokenGS
& 5.1164
& 34.1712
& 2.7911 \\
& LocusGS
& \textbf{0.1978}
& \textbf{0.1371}
& \textbf{0.1117} \\

\addlinespace[2pt]

\multirow{2}{*}{DL3DV}
& TokenGS
& 0.5881
& 0.0139
& 0.5531 \\
& LocusGS
& \textbf{0.0433}
& \textbf{0.0004}
& \textbf{0.0383} \\

\bottomrule
\end{tabular}
\end{table}

\subsection{In-depth Analysis}

% Analysis 2:
\subsubsection{Gaussian Distribution}
We further visualize the global spatial organization of the predicted
Gaussian representation in \Cref{fig:global_gs_compare}. Given the
same input views, TokenGS tends to produce more diffuse Gaussian clouds
with scattered or floating primitives, whereas LocusGS yields a more
structured distribution that better follows the main scene geometry. In
addition to the final Gaussians, we visualize the learned anchors of
LocusGS. The anchors form a coarse spatial scaffold over the scene,
providing explicit geometric support for token specialization and
subsequent Gaussian decoding. This indicates that LocusGS improves not
only rendering quality, but also the spatial organization and
interpretability of token-based 3DGS representations.

\subsubsection{Anchor Distribution and Adaptive Radii}
We analyze the spatial distribution of the learned anchors and their
adaptive radii. For visualization, we project the anchors from the last
decoder layer onto each input view. The first row of
\Cref{fig:token_compactness} shows all projected anchors, while the
second row highlights several selected anchors together with their
projected radii. The third row further visualizes the Gaussian groups
decoded from these selected anchors, with different colors indicating
different anchors.
As shown, the learned anchors cover
the main scene structures across different views, suggesting that they
form a coarse spatial scaffold for token-based Gaussian prediction. Their
distribution is also adaptive rather than uniform: anchors tend to
concentrate around visually or geometrically informative regions, while
less constrained regions are covered more sparsely. The learned radii
show a similar adaptive behavior. Anchors in sparse or weakly constrained
areas often have larger radii, providing broader spatial support, whereas
anchors around detailed regions use smaller radii to focus on local
structures. This indicates that LocusGS learns not only anchor locations,
but also meaningful support scales for organizing token-associated
Gaussians.

\subsubsection{Token-level Spatial Compactness}
We examine the spatial organization of the Gaussian groups decoded from
individual tokens. In \Cref{fig:token_compactness}, different colors
indicate Gaussians generated by different tokens. LocusGS produces
compact local groups around the corresponding anchors, whereas TokenGS
often generates more spatially scattered groups.
%  Since the two models
% learn their tokens independently, the visualization compares
% representative token-level behaviors rather than establishing
% one-to-one token correspondence.

We further quantify this property using token-level Gaussian dispersion.
For token \(i\), let
\(\{\boldsymbol{\mu}^{G}_{i,k}\}_{k=1}^{K}\) denote the centers of its
\(K\) decoded Gaussians. We define
\begin{equation}
\begin{aligned}
\bar{\boldsymbol{\mu}}^{G}_{i}
&=
\frac{1}{K}
\sum_{k=1}^{K}
\boldsymbol{\mu}^{G}_{i,k},
\\
C_{\mathrm{centroid}}
&=
\frac{1}{NK}
\sum_{i=1}^{N}
\sum_{k=1}^{K}
\left\|
\boldsymbol{\mu}^{G}_{i,k}
-
\bar{\boldsymbol{\mu}}^{G}_{i}
\right\|_2
\end{aligned}
\end{equation}
where \(N\) is the number of Gaussian tokens. Lower values indicate
tighter within-token Gaussian groups. Since RE10K and DL3DV use
different scene normalization conventions, values are only compared
between methods within the same dataset.
As shown in \Cref{tab:quan_token_compactness}, LocusGS substantially
reduces token-level Gaussian dispersion on both datasets. Together with
the improved rendering quality, these results show that the proposed
anchor-based formulation yields more coherent local Gaussian groups
without compromising reconstruction accuracy.

\begin{figure*}[t]
    \centering
    \includegraphics[width=\textwidth]{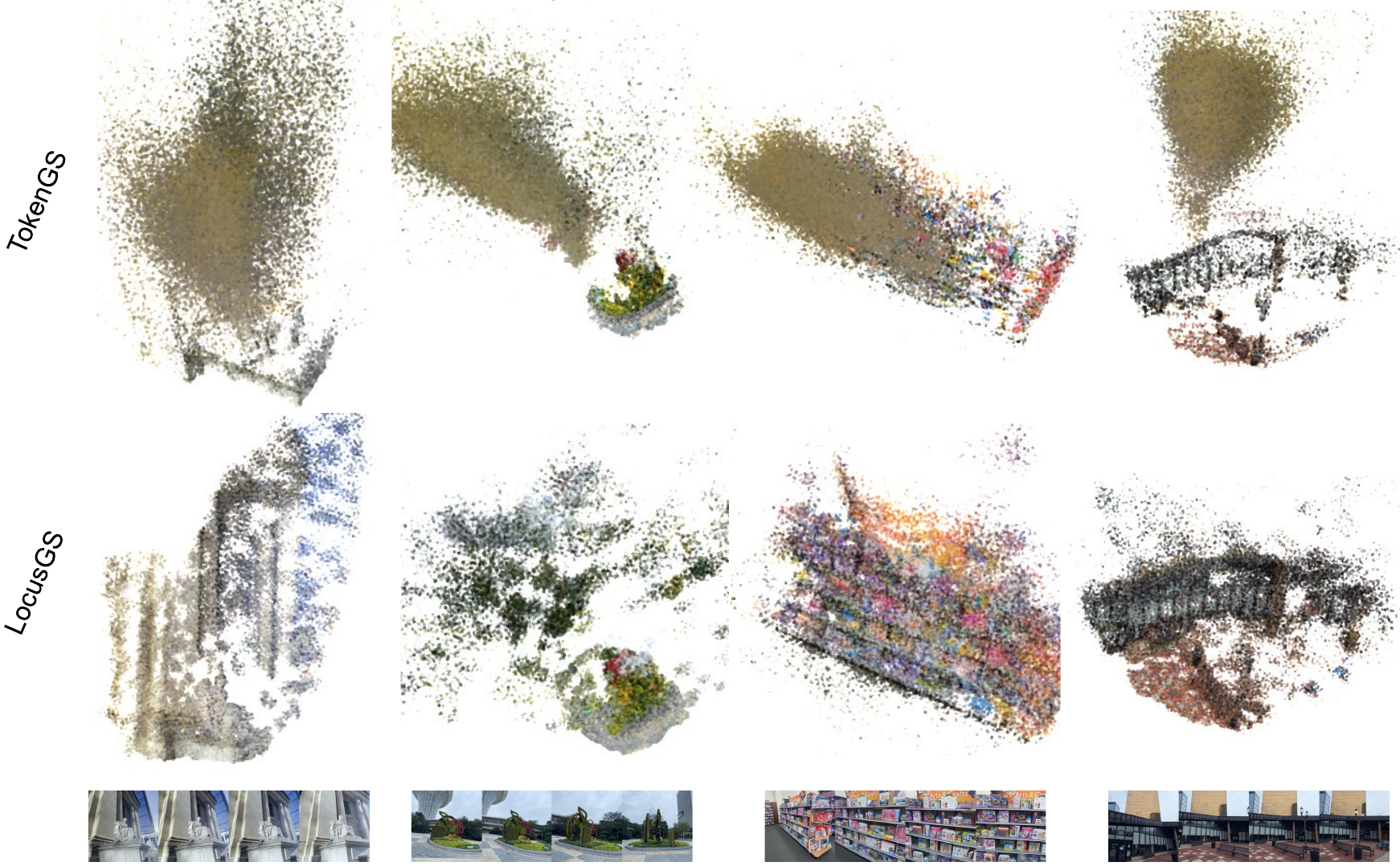}
    \caption{
        \textbf{Qualitative comparisons of the reconstructed Gaussians produced by TokenGS and LocusGS.}
        }
    \label{fig:more_3dgs}
\end{figure*}

\subsubsection{Cross Attention Visualization}
\begin{figure}[t!]
    \centering
    \includegraphics[width=0.7\columnwidth]{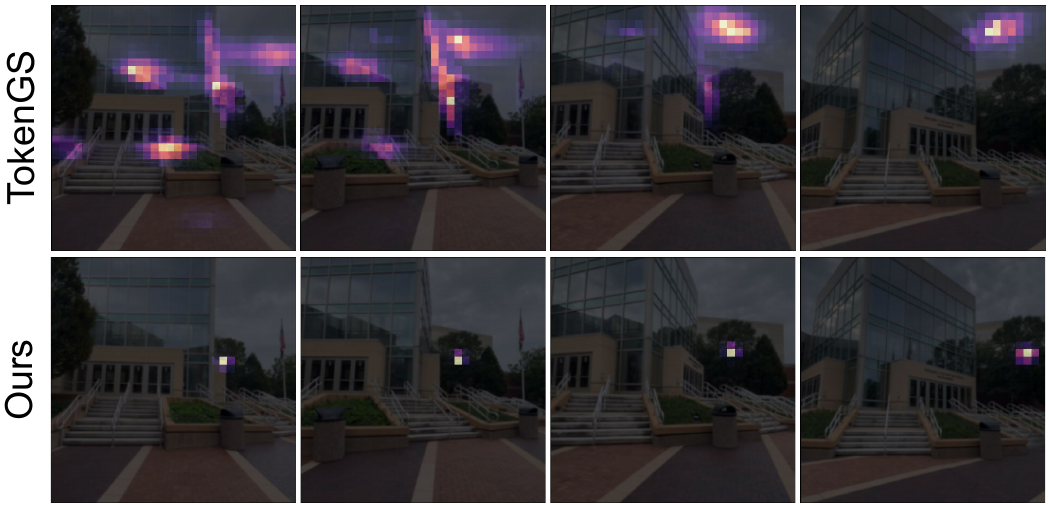}
    \caption{
        \textbf{Spatial anchors yield more localized and view-consistent cross-view attention.}
        For each method, we show the cross-attention map of a representative token.
        Our method produces more localized and view-consistent attention patterns around the projected anchor region.
        }
    \label{fig:attention_analysis}
\end{figure}

We visualize the decoder cross-attention in the last decoder layer to examine how tokens aggregate image evidence across views. 
For each selected token, we average the cross-attention weights over all heads, reshape the resulting attention vector into per-view patch grids, and overlay the attention maps on the corresponding input images. 
We also project the Gaussian centers decoded from the same token onto each input view using the camera intrinsics and extrinsics. 
As shown in \Cref{fig:attention_analysis}, TokenGS produces scattered attention responses over multiple image regions, while LocusGS shows more localized and view-consistent attention around its projected anchor and decoded Gaussian centers. 
This indicates that the explicit anchor state provides a geometric prior for cross-view feature aggregation, encouraging each token to gather evidence from geometrically relevant image regions rather than relying solely on content-based similarity. 
Because tokens may specialize to different scene regions, their indices
are not comparable between TokenGS and LocusGS.
We therefore compare representative attention patterns rather than
one-to-one token correspondences.

\subsubsection{Training Curves}
\begin{figure*}[t]
    \centering
    \includegraphics[width=0.7\textwidth]{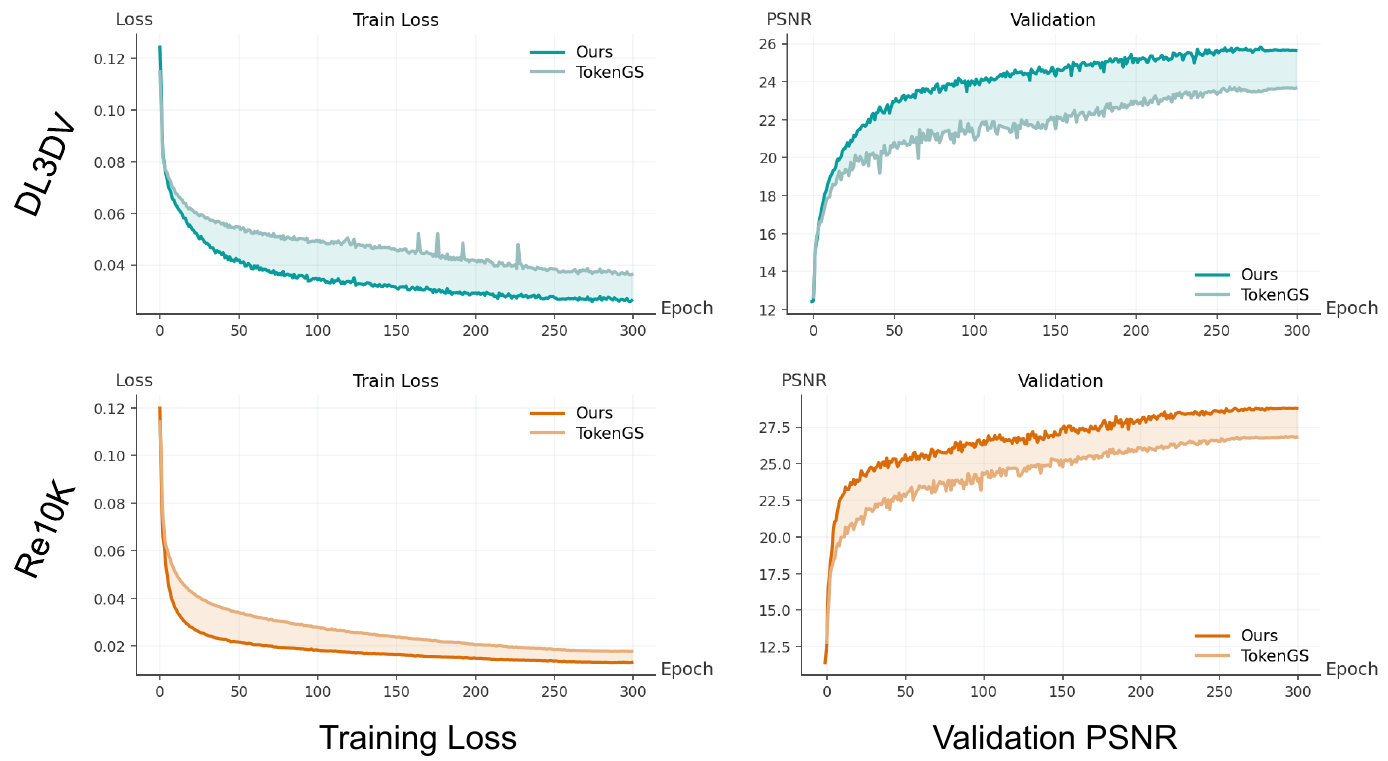}
    \caption{
        \textbf{Convergence comparison on DL3DV and RE10K.}
        Our method exhibits faster convergence with respect to training epochs, reaches lower training loss, and achieves consistently higher PSNR on both training and validation sets.
        }
    \label{fig:training_curves}
\end{figure*}

\Cref{fig:training_curves} compares the training dynamics of TokenGS and our method.
Our model exhibits faster convergence with respect to training epochs and maintains a clear advantage in both training and validation PSNR, suggesting that the anchor-guided formulation provides a more effective optimization path for token-based Gaussian prediction.

\subsubsection{Decomposition of anchor-guided cross-attention.}
\begin{figure}[t]
    \centering
    \includegraphics[width=0.7\columnwidth]{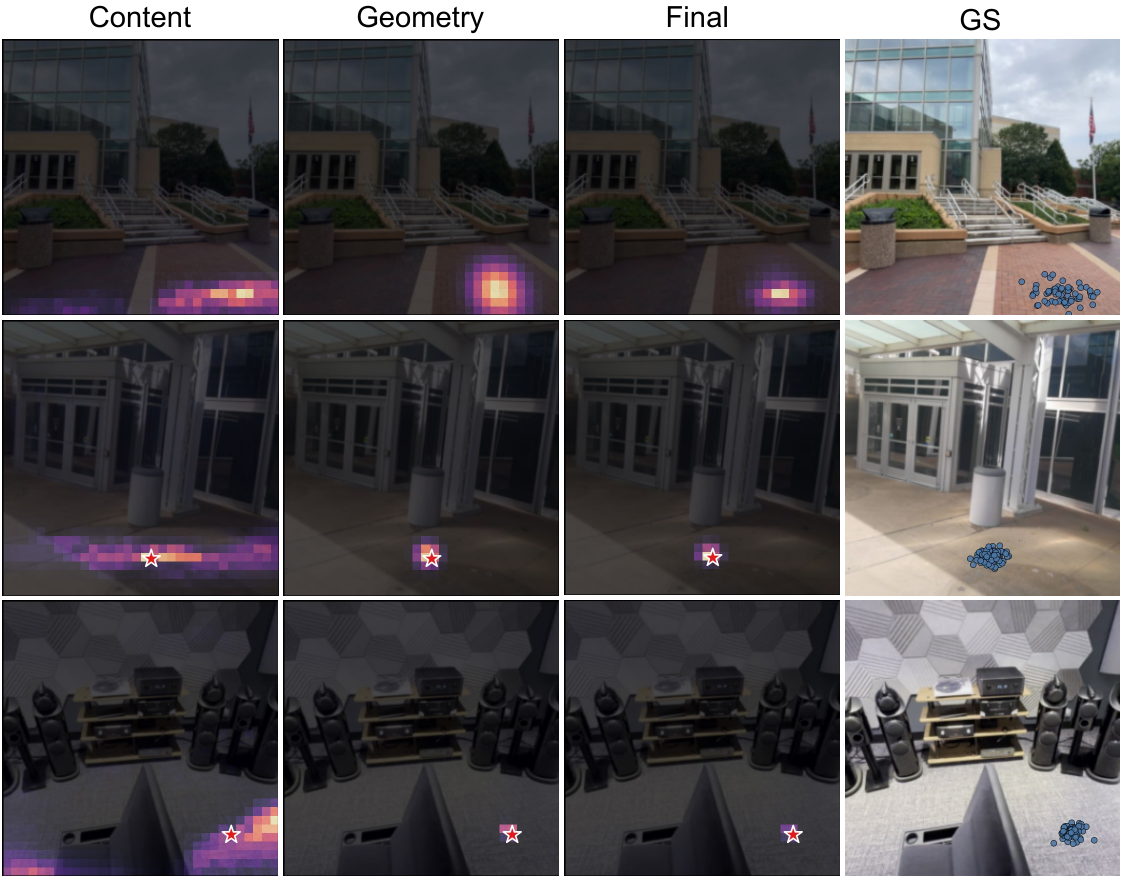}
    \caption{
        \textbf{Decomposition of anchor-guided cross-attention.}
        We visualize the content-only attention, geometry-only attention, final attention, and projected Gaussians from the same token.
        The final attention combines visual similarity with geometric compatibility, leading to more localized evidence aggregation.
        }
    \label{fig:attention_decomposition}
\end{figure}

To further inspect how the geometric prior affects cross-view feature aggregation, we decompose the cross-attention logits in the last decoder layer into the content term and the geometry term. 
The content term is computed from the standard query-key similarity, while the geometry term is given by the point-to-ray bias between the current 3D anchor and the Pl\"ucker rays of image patches. 
For visualization, we separately normalize the content logits and geometry logits with softmax, and compare them with the final attention obtained by applying softmax after adding the two terms. 
We also project the Gaussian centers decoded from the same token onto the input view. 
As shown in \Cref{fig:attention_decomposition}, the content branch may respond to broad or ambiguous image regions with similar appearance, whereas the geometry branch provides a localized spatial prior around regions consistent with the anchor position. 
The final attention combines these two cues and focuses on image evidence that is both visually relevant and geometrically plausible. 
This supports our design motivation that anchor-guided cross-attention reduces purely appearance-driven ambiguity and encourages each token to aggregate information from cross-view regions consistent with its 3D spatial hypothesis.

\begin{table*}[t]
\centering
\caption{
Ablation of the structural components of LocusGS under the 4-view setting.
All variants use the $256\times256$ base model. 
% without the $448\times256$ finetuning used for the main results in
% \Cref{tab:dl3dv_view_generalization}.
Checkmarks denote enabled components, while ``---'' denotes absent or
inapplicable components.
}
\label{tab:LocusGS_structural_ablation}
\footnotesize
\renewcommand{\arraystretch}{1.15}
\setlength{\tabcolsep}{4pt}

\resizebox{\textwidth}{!}{
\begin{tabular}{llcccccccccc}
\toprule

\multirow{2.4}{*}{Ablation Group}
&
\multirow{2.4}{*}{Variant}
&
\multicolumn{2}{c}{Anchor State}
&
\multicolumn{2}{c}{Layer-wise Refinement}
&
\multirow{2.4}{*}{\makecell{Anchor-Aware\\Self-Attn.}}
&
\multirow{2.4}{*}{\makecell{Cross-Attn.\\Bias}}
&
\multirow{2.4}{*}{\makecell{Gaussian Center\\Decoding}}
&
\multirow{2.4}{*}{PSNR$\uparrow$}
&
\multirow{2.4}{*}{SSIM$\uparrow$}
&
\multirow{2.4}{*}{LPIPS$\downarrow$}
\\

\cmidrule(lr){3-4}
\cmidrule(lr){5-6}
\noalign{
    \vskip-\dimexpr\belowrulesep-\aboverulesep\relax
}

&
&
Center
&
Radius
&
Center
&
Radius
&
&
&
&
&
&
\\

\midrule

% \rowcolor{gray!10}
\textit{Reference}
&
\textbf{Full model}
&
\checkmark
&
\checkmark
&
\checkmark
&
\checkmark
&
\checkmark
&
Center + Radius
&
Radius-scaled offset
&
\textbf{24.284}
&
\textbf{0.7843}
&
\textbf{0.2709}
\\

\midrule

\multirow{2}{*}{
\makecell[l]{\textit{Anchor}\\\textit{Representation}}
}
&
No radius state
&
\checkmark
&
\textemdash
&
\checkmark
&
\textemdash
&
\checkmark
&
Center
&
Anchor offset
&
23.640
&
0.7618
&
0.3007
\\

&
Static radius w/o refinement
&
\checkmark
&
\checkmark
&
\checkmark
&
\textemdash
&
\checkmark
&
Center + Radius
&
Radius-scaled offset
&
23.979
&
0.7746
&
0.2866
\\

\midrule

\multirow{3}{*}{
\makecell[l]{\textit{Attention}\\\textit{Design}}
}
&
Content-only self-attention
&
\checkmark
&
\checkmark
&
\checkmark
&
\checkmark
&
\textemdash
&
Center + Radius
&
Radius-scaled offset
&
24.171
&
0.7807
&
0.2764
\\

&
Content-only cross-attention
&
\checkmark
&
\checkmark
&
\checkmark
&
\checkmark
&
\checkmark
&
Content
&
Radius-scaled offset
&
22.751
&
0.7220
&
0.3500
\\

&
Center-only geometric bias
&
\checkmark
&
\checkmark
&
\checkmark
&
\checkmark
&
\checkmark
&
Center
&
Radius-scaled offset
&
23.770
&
0.7670
&
0.2940
\\

\midrule

\multirow{1}{*}{
\makecell[l]{\textit{Gaussian Decoding}}
}
&
Free Gaussian centers
&
\checkmark
&
\checkmark
&
\checkmark
&
\checkmark
&
\checkmark
&
Center + Radius
&
Free center
&
22.079
&
0.6860
&
0.3970
\\

\bottomrule
\end{tabular}
}
\end{table*}

\subsubsection{Self Attention Visualization}
\begin{figure}[t]
    \centering
    \includegraphics[width=0.6\columnwidth]{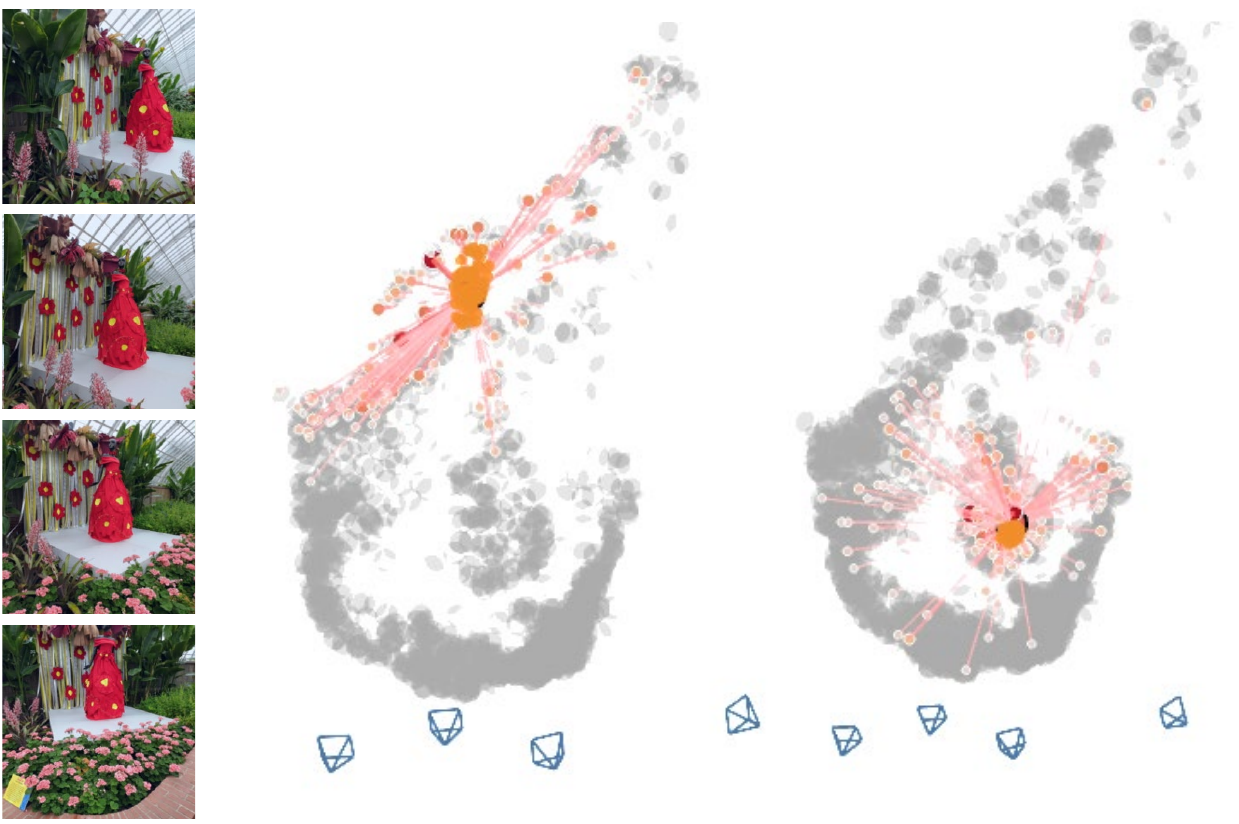}
    \caption{
        \textbf{Visualization of anchor-aware self-attention.}
        For each selected query anchor, we visualize its decoded Gaussian group and the top-k attended anchors. Edge thickness and opacity indicate attention strength.
        }
    \label{fig:self_attn_vis}
\end{figure}

We further visualize the self-attention behavior among anchor tokens.
Different from cross-attention, which associates anchor tokens with
multi-view image observations, self-attention models the interaction
between Gaussian tokens themselves. 
To inspect this interaction, we select a query anchor and visualize the
anchors that receive the highest self-attention weights from it.
For each selected query anchor, we highlight the query anchor in red and
keep its decoded Gaussian group visible. 
We then retrieve the top-\(k\) anchors according to the self-attention
weights of the query anchor. 
These attended anchors are visualized with color and size proportional to
their attention weights, and we draw edges from the query anchor to the
attended anchors. 
Thicker and more opaque edges indicate stronger self-attention weights.
The remaining Gaussians are shown in gray to provide the global scene
context.
As shown in \Cref{fig:self_attn_vis}, a query anchor mainly interacts
with a sparse set of other anchors rather than uniformly attending to all
tokens. 
These attended anchors are not merely a visualization of the full anchor
layout; instead, they reveal the information-exchange neighborhood of the
selected token. 
This suggests that anchor-aware self-attention allows each spatial token
to aggregate contextual information from related anchors.

\subsection{Ablation Studies}

\Cref{tab:LocusGS_structural_ablation} evaluates the key components of
LocusGS on DL3DV under the 4-view setting.
Modeling both the anchor center and radius is beneficial: removing the
radius or keeping it static consistently degrades performance, showing
that the spatial support should adapt throughout decoding.
Anchor-aware self-attention also improves over content-only
self-attention, indicating that explicit anchor positions provide useful
spatial cues for token interactions.
Anchor-aware cross-attention is more important, as removing the
anchor-to-ray bias leads to a substantial degradation, while the
radius-adaptive bias further improves over its center-only counterpart.
The largest performance drop occurs when Gaussian centers are predicted
freely, confirming that anchor-centered, radius-scaled decoding is
crucial for maintaining the spatial association between each token and
its Gaussian group.
\Cref{tab:multi_layer_supervision_ablation} further examines intermediate
rendering supervision.
Supervising only the final decoder layer performs clearly worse, whereas
adding supervision at the middle layer yields the best results.
A denser supervision schedule provides no further improvement,
indicating that supervision at the middle and final layers is sufficient.

\begin{table}[t]
\centering
\caption{
Ablation of multi-layer rendering supervision.
%  under the 4-view setting.
$\mathcal{S}$ denotes the set of decoder layers receiving rendering supervision.
% All variants are evaluated using the base model at
% $256\times256$ resolution.
}
\label{tab:multi_layer_supervision_ablation}
\footnotesize
\renewcommand{\arraystretch}{1.12}
\begin{tabular}{lccc}
\toprule
Supervised Layers $\mathcal{S}$ & PSNR$\uparrow$ & SSIM$\uparrow$ & LPIPS$\downarrow$ \\
\midrule
$\{6,12\}$ & \textbf{24.284} & \textbf{0.7843} & \textbf{0.2709} \\
$\{12\}$ & 23.549 & 0.7575 & 0.3074 \\
$\{3,6,9,12\}$ & 24.071 & 0.7793 & 0.2774 \\
\bottomrule
\end{tabular}
\end{table}

\section{Conclusion}

We presented LocusGS, a spatially grounded anchor-token formulation for
feed-forward 3D Gaussian reconstruction. By augmenting each Gaussian
token with an explicit 3D anchor state and refining it across decoder
layers, LocusGS provides spatial references for cross-view aggregation
and anchor-centered Gaussian decoding. This design improves rendering
quality while producing more coherent Gaussian distributions, compact
token-level Gaussian groups, and meaningful anchor layouts. Our results
indicate that explicit anchor states offer an effective spatial prior for
query-based feed-forward 3DGS, turning Gaussian tokens from implicit
latent embeddings into more interpretable spatial reconstruction queries.

\clearpage

\appendix
{

\section{Implementation Details}

\subsection{Overview of the Encoder--Decoder Framework}

LocusGS follows an encoder--decoder framework similar to query-based feed-forward methods~\cite{tokengs2026}. 
Given posed multi-view images, we first tokenize each input image into patch-level visual tokens using a shared image patch embedding layer. 
Camera information is injected by adding a Pl\"ucker-ray embedding to the corresponding image patch features. 
The resulting multi-view image tokens are concatenated across views and processed by the image encoder to produce the keys and values for decoder cross-attention.

The decoder maintains a fixed set of learnable Gaussian tokens. 
Each token is associated with an explicit anchor state consisting of a 3D anchor center and a scalar support radius. 
Each decoder layer updates the Gaussian token features through anchor-guided cross-attention, anchor-aware self-attention, and a feed-forward network. 
After selected decoder layers, the current tokens and anchor states are decoded into Gaussian primitives.

Unless otherwise stated, we use the same token budget as the corresponding TokenGS baseline, while fixing the number of Gaussians predicted per token to 64.
As a result, each token corresponds to one Gaussian group containing 64 Gaussians.
This gives 262{,}144 Gaussians in total for the 4096-token setting and 65{,}536 Gaussians in total for the 1024-token setting.

We train the model with AdamW~\cite{loshchilov2017decoupled} and a cosine learning rate schedule, using an initial learning rate of $4 \times 10^{-4}$ with $2{,}000$ warmup iterations for base training, and $4 \times 10^{-5}$ with $400$ warmup iterations for finetuning. 
We evaluate novel view synthesis quality using PSNR, SSIM, and LPIPS.

\subsection{Choice of Query-Based Baseline}
Several concurrent methods adopt token-based Gaussian prediction but differ in their input assumptions and system components, including whether camera poses are known and whether pretrained geometric reconstruction backbones are used. Despite these differences, they share a common abstraction: latent queries aggregate image evidence and are subsequently decoded into Gaussian primitives~\cite{globalsplat,tokensplat,an2025c3g}. We choose TokenGS as our primary query-based baseline because it provides the cleanest controlled instantiation of this abstraction. Specifically, TokenGS uses posed inputs, learnable Gaussian queries, and no additional pretrained geometric reconstruction backbone, allowing us to match both the token count and Gaussian budget. This controlled setting enables us to isolate the effect of replacing implicit query embeddings with explicit, dynamically refined spatial states. In contrast, other concurrent token-based systems introduce additional variations in pose estimation, feature extraction, or representation capacity~\cite{an2025c3g,globalsplat}, which would confound a direct assessment of spatial grounding. TokenGS therefore serves not as a proxy for the complete systems of all concurrent methods, but as a canonical implementation of their shared query-based Gaussian prediction core and the most appropriate baseline for our controlled evaluation.

\subsection{Radius Parameterization and Usage}

Each anchor is associated with a scalar support radius, which controls the spatial extent of the corresponding token-associated Gaussian group.
In implementation, we distinguish between the \emph{raw radius} and the \emph{activated radius}. 
The raw radius $\rho_i \in \mathbb{R}$ is an unconstrained learnable parameter, while the activated radius $r_i \in \mathbb{R}^{+}$ is the positive support scale used by the model.

Specifically, the activated radius is obtained by
\begin{equation}
r_i^l
=
\operatorname{softplus}(\rho_i^l)
+
\epsilon
\end{equation}
where \(\epsilon>0\) is a small constant for numerical stability.
We initialize the raw radius by applying the inverse softplus function to the desired initial radius, so that the activated radius starts from the predefined support scale.
In our default setting, each Gaussian token has its own independent scalar radius parameter.
% We set the initial activated radius to $1.0$ and use a lower bound $r_{\min}=10^{-3}$.

During decoding, we update the raw radius rather than the activated radius. 
After the $l$-th decoder layer, the updated token feature predicts a residual update:
\begin{equation}
    \Delta \rho_i^l = f_r(\mathbf{q}^{l+1}_i),
\end{equation}
and the raw radius is refined as
\begin{equation}
    \rho_i^{l+1} = \rho_i^l + \Delta \rho_i^l .
\end{equation}
The activated radius used at the next layer is then computed as
\begin{equation}
r_i^{l+1}
=
\operatorname{softplus}(\rho_i^{l+1})
+
\epsilon
\end{equation}
This parameterization allows the radius to be optimized freely in an unconstrained space while ensuring that the actual support scale remains positive.

The activated radius is used in two places. 
First, in radius-adaptive anchor-to-ray attention, it controls the bandwidth of the geometric bias:
\begin{equation}
    \sigma_i^l = \sigma_0 r_i^l ,
\end{equation}
where $\sigma_0$ is a fixed base bandwidth. 
In our implementation, we use a fixed base bandwidth $\sigma_0 = 0.1$.
A larger activated radius increases the bandwidth of the geometric bias, making the geometry-induced penalty smoother and less selective with respect to point-to-ray distance. 
Conversely, a smaller radius produces a sharper geometric bias that more strongly suppresses rays far from the anchor.

Second, in anchor-centered Gaussian decoding, the activated radius scales the local Gaussian offsets:
\begin{equation}
    \boldsymbol{\mu}^{G}_{i,k}
    =
    \boldsymbol{\mu}^{l}_{i}
    +
    r_i^l \boldsymbol{\delta}_{i,k}.
\end{equation}
Here, $\boldsymbol{\mu}^{l}_{i}$ is the anchor center and $\boldsymbol{\delta}_{i,k}$ is the predicted local offset for the $k$-th Gaussian decoded from token $i$. 
Thus, the radius acts as the local coordinate scale of each token-associated Gaussian group.
By default, each Gaussian token predicts a group of 64 Gaussians.

In summary, the raw radius $\rho_i$ is used for optimization and residual refinement, whereas the activated radius $r_i$ is used as the physical support scale for geometric attention and Gaussian decoding.

\begin{table}[t]
\centering
\caption{
Formal definitions of the cross-attention and Gaussian center decoding
variants used in the ablation study.
}
\label{tab:ablation_variant_definitions}
\footnotesize
\renewcommand{\arraystretch}{1.35}
\begin{tabular}{lll}
\toprule
Category & Variant & Formal definition \\
\midrule

\multirow{3}{*}[-1.0ex]{Cross-Attention}
&
Content only
&
\(\displaystyle
\alpha_{ij}^{\,l}
=
\operatorname{softmax}_{j}
\left(
c_{ij}^{\,l}
\right)
\)
\\[0.3em]

&
Center-only
&
\(\displaystyle
b_{ij}^{\,l}
=
-\frac{1}{2}
\left(
\frac{D_{ij}^{\,l}}{\sigma_0}
\right)^2,
\quad
\alpha_{ij}^{\,l}
=
\operatorname{softmax}_{j}
\left(
c_{ij}^{\,l}
+
\gamma b_{ij}^{\,l}
\right)
\)
\\[0.3em]

&
Radius-adaptive
&
\(\displaystyle
b_{ij}^{\,l}
=
-\frac{1}{2}
\left(
\frac{D_{ij}^{\,l}}{\sigma_0 r_i^{\,l}}
\right)^2,
\quad
\alpha_{ij}^{\,l}
=
\operatorname{softmax}_{j}
\left(
c_{ij}^{\,l}
+
\gamma b_{ij}^{\,l}
\right)
\)
\\

\midrule

\multirow{3}{*}[-0.4ex]{Gaussian Decoding}
&
Free center
&
\(\displaystyle
\boldsymbol{\mu}_{i,k}^{G,l}
=
\boldsymbol{\delta}_{i,k}^{\,l}
\)
\\[0.2em]

% &
% Anchor offset
% &
% \(\displaystyle
% \boldsymbol{\mu}_{i,k}^{G,l}
% =
% \boldsymbol{\mu}_{i}^{\,l}
% +
% \boldsymbol{\delta}_{i,k}^{\,l}
% \)
% \\[0.2em]

&
Radius-scaled offset
&
\(\displaystyle
\boldsymbol{\mu}_{i,k}^{G,l}
=
\boldsymbol{\mu}_{i}^{\,l}
+
r_i^{\,l}
\boldsymbol{\delta}_{i,k}^{\,l}
\)
\\

\bottomrule
\end{tabular}
\end{table}

\begin{table}[t]
  \centering
  \caption{
  Training and evaluation settings used in our experiments.
  We report the input-view setting, resolution, token budget, and optimization schedule for each dataset.
  }
  \label{tab:training_protocol}
%   \footnotesize
%   \renewcommand{\arraystretch}{1.15}
  \begin{tabular}{lcc}
  \toprule
   & RE10K & DL3DV \\
  \midrule
  Training input views & 2 & 4 \\
  Evaluation input views & 2 & 2 / 4 / 6 \\
  Base training resolution & $256 \times 256$ & $256 \times 256$ \\
  Finetuning / evaluation resolution & $256 \times 256$ & $448 \times 256$ \\
  % Encoder / decoder layers & -- / -- & 3 / 12 \\
  Gaussian tokens & 1024 base, 4096 finetune & 1024 base, 4096 finetune \\
  Gaussians per token & 64 & 64 \\
  Optimizer & AdamW & AdamW \\
  Base learning rate & $4 \times 10^{-4}$ & $4 \times 10^{-4}$ \\
  Finetuning learning rate & $4 \times 10^{-5}$ & $4 \times 10^{-5}$ \\
  Base / finetuning warmup & 2000 / 400 & 2000 / 400 \\
  Base / finetuning epochs & 300 / 20 & 300 / 20 \\
  \bottomrule
  \end{tabular}
\end{table}

\subsection{Patch-level Pl\"ucker Ray Construction}

Camera geometry is represented using Pl\"ucker rays associated with image patches. 
For each input view, we first construct a dense pixel-level Pl\"ucker ray map from the camera intrinsics and extrinsics. 
Each ray is represented as
\begin{equation}
    \ell = (\mathbf{m}, \mathbf{d}) \in \mathbb{R}^{6},
\end{equation}
where $\mathbf{d}$ is the ray direction and $\mathbf{m}$ is the moment vector. 
We use the convention
\begin{equation}
    \mathbf{m} = \mathbf{o} \times \mathbf{d},
\end{equation}
where $\mathbf{o}$ denotes the camera center in the scene coordinate system.

Since the image encoder operates on patch tokens, we convert the dense Pl\"ucker map into patch-level ray features. 
Given a Pl\"ucker tensor of shape $B \times V \times 6 \times H \times W$, where $B$ is the batch size and $V$ is the number of input views, we apply average pooling with the same kernel size and stride as the image patch size $P$. 
This produces a patch-level Pl\"ucker tensor:
\begin{equation}
    \mathbf{L}_{\mathrm{patch}}
    \in
    \mathbb{R}^{B \times (V H_p W_p) \times 6},
    \qquad
    H_p = H / P,\quad W_p = W / P .
\end{equation}
Each patch-level ray therefore provides an approximate geometric representation for the corresponding image token.

\subsection{Point-to-Ray Distance in Pl\"ucker Coordinates}

In the main paper, the geometric bias is defined using the shortest
Euclidean distance \(D(\boldsymbol{\mu}_i^{\,l}, \boldsymbol{\ell}_j)\)
from the current anchor center to an image ray. 
Here we provide the concrete computation used in our implementation.

Each patch-level camera ray is represented by Pl\"ucker coordinates
\(\boldsymbol{\ell}_j=(\mathbf{m}_j,\mathbf{d}_j)\), where
\(\mathbf{d}_j\) is the ray direction and \(\mathbf{m}_j\) is the moment
vector. We use the convention
\begin{equation}
    \mathbf{m}_j = \mathbf{o}_j \times \mathbf{d}_j ,
\end{equation}
where \(\mathbf{o}_j\) is the camera center. Before computing the
distance, we normalize the ray direction and scale the moment
accordingly:
\begin{equation}
    \bar{\mathbf{d}}_j =
    \frac{\mathbf{d}_j}{\|\mathbf{d}_j\|_2},
    \qquad
    \bar{\mathbf{m}}_j =
    \frac{\mathbf{m}_j}{\|\mathbf{d}_j\|_2}.
\end{equation}
Under this convention, a 3D point \(\boldsymbol{\mu}\) lies on the ray if
\(\boldsymbol{\mu}\times\bar{\mathbf{d}}_j=\bar{\mathbf{m}}_j\).
Therefore, we compute the point-to-ray distance as
\begin{equation}
    D(\boldsymbol{\mu}_i^{\,l}, \boldsymbol{\ell}_j)
    =
    \left\|
    \boldsymbol{\mu}_i^{\,l}
    \times
    \bar{\mathbf{d}}_j
    -
    \bar{\mathbf{m}}_j
    \right\|_2 .
\end{equation}

This distance is then substituted into the geometric bias in the main
paper:
\begin{equation}
b_{ij}^{\,l}
=
-
\tfrac{1}{2}
\left(
\frac{
D\left(\boldsymbol{\mu}_i^{\,l}, \boldsymbol{\ell}_j\right)
}{
\sigma_0 r_i^{\,l}
}
\right)^2 .
\end{equation}
The activated radius \(r_i^{\,l}\) modulates the bandwidth of this
geometric prior. 
In our default setting, the base bandwidth is fixed to \(\sigma_0 = 0.1\).
For stability, the squared bandwidth is lower-bounded.
The geometric bias is clamped to the interval \([-20, 0]\) to avoid overly sharp attention scores.
The final cross-attention weights still follow the
formulation in the main paper and are jointly determined by content
similarity, the geometric bias, and the learnable scale \(\gamma\).

\subsection{Training Objective}

We follow the rendering objective of TokenGS~\cite{tokengs2026}. 
For each supervised decoder layer \(l_m \in \mathcal{S}\), we decode the
current tokens and anchors into a Gaussian set \(\mathcal{G}^{l_m}\),
render it to the target view, and apply the same image reconstruction
loss as the baseline:
\begin{equation}
    \mathcal{L}_{\mathrm{rec}}^{\,l_m}
    =
    \mathcal{L}_{\mathrm{MSE}}^{\,l_m}
    +
    \lambda_{\mathrm{SSIM}}
    \mathcal{L}_{\mathrm{SSIM}}^{\,l_m}
\end{equation}

For training stability, we also adopt the visibility regularization used
in TokenGS. Given a set of 3D points \(\mathcal{X}\), this term softly
penalizes points that are outside all supervision views:
\begin{equation}
\phi(\tilde{u},\tilde{v})
=
\mathrm{ReLU}(|\tilde{u}|-1)
+
\mathrm{ReLU}(|\tilde{v}|-1).
\end{equation}
\begin{equation}
\mathcal{L}_{\mathrm{vis}}(\mathcal{X})
=
\frac{1}{|\mathcal{X}|}
\sum_{\mathbf{x}\in\mathcal{X}}
\min_{\Pi_t\in\Pi_{\mathrm{sup}}}
\phi(\tilde{u}^{\,t},\tilde{v}^{\,t}).
\end{equation}
where \((\tilde{u}^{\,t},\tilde{v}^{\,t})\) are the normalized projected
coordinates of \(\mathbf{x}\) in supervision view \(\Pi_t\). 
Different from the baseline, we apply this regularization to both the
decoded Gaussian centers and the anchor centers:
\begin{equation}
    \mathcal{L}^{l_m}
    =
    \mathcal{L}_{\mathrm{rec}}^{\,l_m}
    +
    \lambda_{\mathrm{G}}
    \mathcal{L}_{\mathrm{vis}}
    \left(
    \{\boldsymbol{\mu}_{i,k}^{G,l_m}\}
    \right)
    +
    \lambda_{\mathrm{A}}
    \mathcal{L}_{\mathrm{vis}}
    \left(
    \{\boldsymbol{\mu}_{i}^{\,l_m}\}
    \right).
\end{equation}
We use $\lambda_{\mathrm{SSIM}}=0.2$, $\lambda_{\mathrm{G}}=1.0$, and $\lambda_{\mathrm{A}}=0.1$ in all experiments.
The final objective is the weighted sum over supervised decoder layers:
\begin{equation}
    \mathcal{L}
    =
    \sum_{m=1}^{M}
    w_m
    \mathcal{L}^{l_m},
    \qquad
    w_m
    =
    \frac{m}{\sum_{n=1}^{M} n}.
\end{equation}

\subsection{Training and Evaluation Protocol}
We also provide the detailed training and evaluation settings in
~\Cref{tab:training_protocol}.

\subsection{Formal Definitions of Ablation Variants}

To clarify the ablation variants in the main paper, we summarize the
formal definitions of the cross-attention and Gaussian center decoding
variants in ~\Cref{tab:ablation_variant_definitions}.
Let
\begin{equation}
    c_{ij}^{\,l}
    =
    \frac{
    \left(\bar{\mathbf{q}}_i^{\,l}\right)^{\top}
    \mathbf{k}_j
    }{
    \sqrt{d}
    }
\end{equation}
denote the content-based attention logit between Gaussian token \(i\)
and image token \(j\) at decoder layer \(l\). 
We use
\(D_{ij}^{\,l}=D(\boldsymbol{\mu}_i^{\,l},\boldsymbol{\ell}_j)\)
for the point-to-ray distance.

\section{Additional Experimental Results}

\subsection{Token-level Spatial Compactness}
\Cref{fig:dl3dv_compactness_score,fig:re10k_compactness_score} compare the per-scene dispersion score of TokenGS and our method on the DL3DV and RE10K 2-view benchmarks. Each bar corresponds to one scene, with the two methods overlaid at the same horizontal position to enable direct comparison.
Since TokenGS generally yields larger dispersion scores, its bar typically appears as the outer, longer bar, while the bar of our method remains shorter and enclosed inside.
A consistent trend can be observed across both datasets: our method achieves substantially lower dispersion scores for nearly all scenes. Since a lower dispersion score indicates that the Gaussians associated with each token or anchor are spatially
more concentrated, these results suggest that our method learns more localized and spatially coherent groupings. In contrast, the larger scores of TokenGS indicate that its token-associated Gaussian groups are more spatially dispersed, implying a
weaker correspondence between the learned representation and the underlying local 3D structure.
Overall, these results support our claim that explicitly grounded 3D anchors lead to more compact and interpretable local representations than token designs without explicit spatial grounding.

\begin{figure*}[t]
    \centering
    \includegraphics[width=\textwidth]{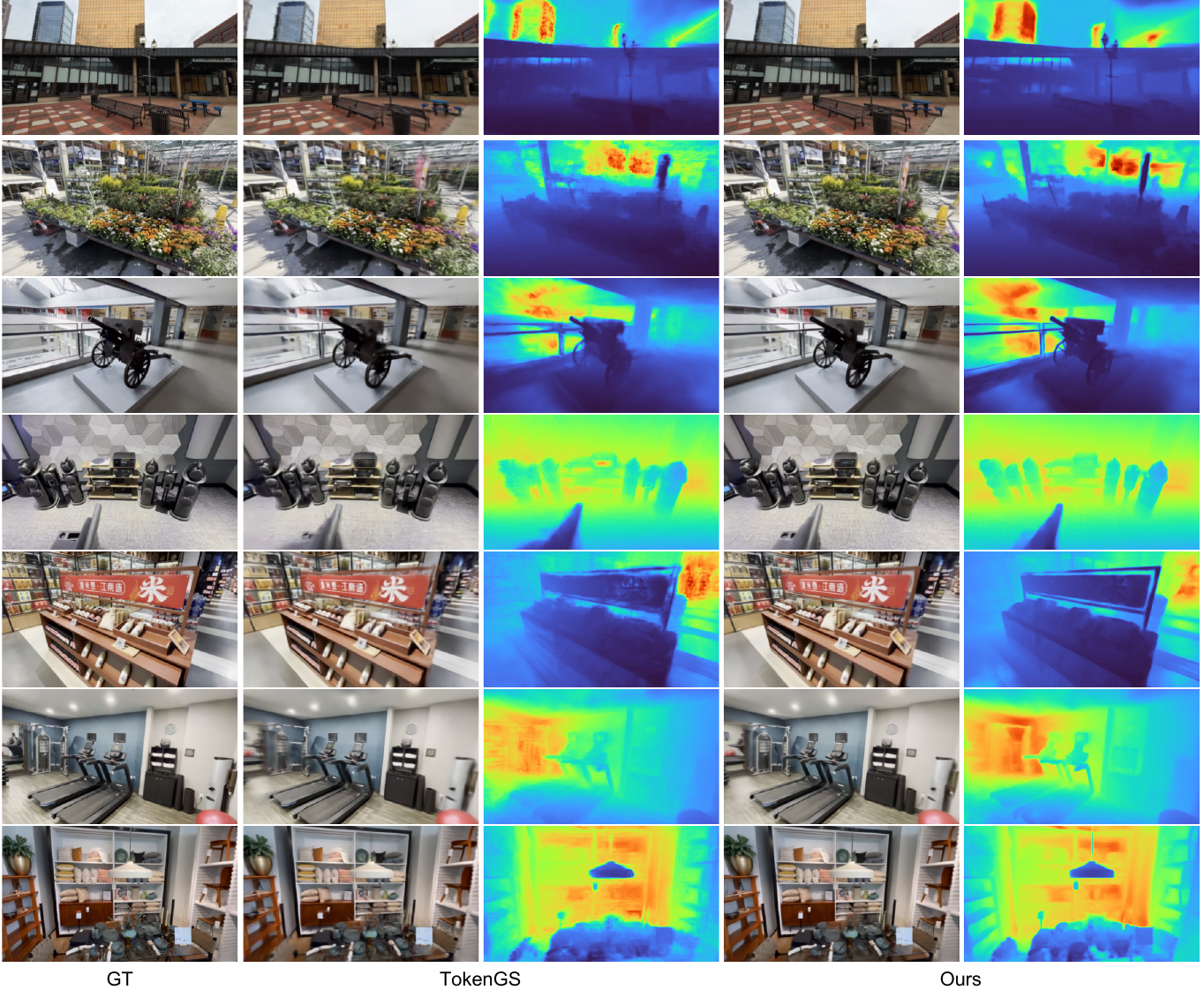}
    \caption{
        \textbf{Qualitative comparison on DL3DV novel-view synthesis.}
        LocusGS produces sharper renderings and more coherent depth structures, especially in texture-rich regions and cluttered scenes.
        }
    \label{fig:nvs_and_depth_more_results}
\end{figure*}

\begin{figure*}[t]
    \centering
    \includegraphics[width=\textwidth]{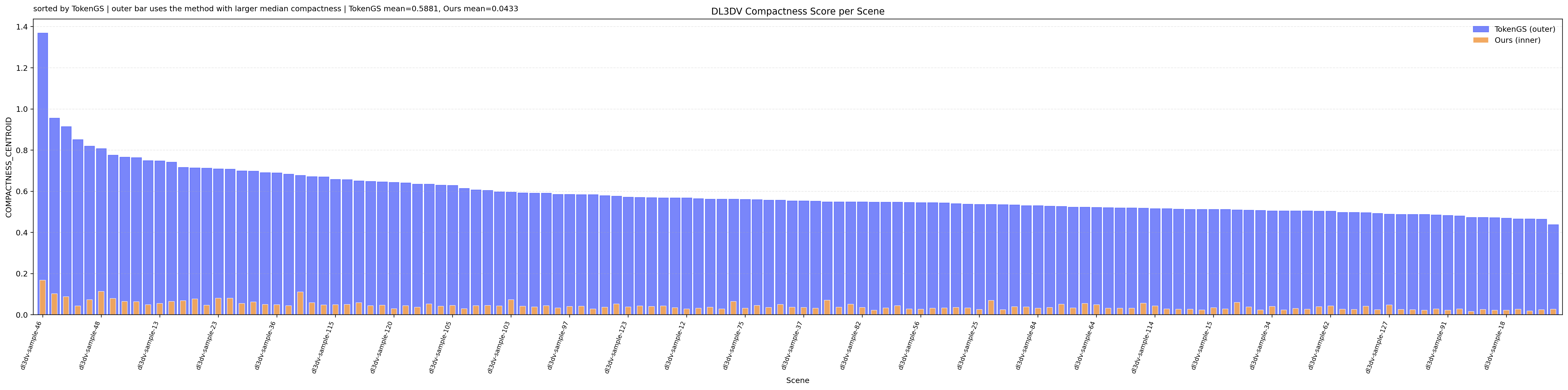}
    \caption{
        \textbf{ Per-scene dispersion score comparison between TokenGS and our method on DL3DV.}
        Each overlaid pair of bars corresponds to the same scene. Our method consistently yields substantially lower
         dispersion scores than TokenGS, indicating that the Gaussian groups associated with each anchor are spatially more compact and coherent. 
        }
    \label{fig:dl3dv_compactness_score}
\end{figure*}
\begin{figure*}[t]
    \centering
    \includegraphics[width=\textwidth]{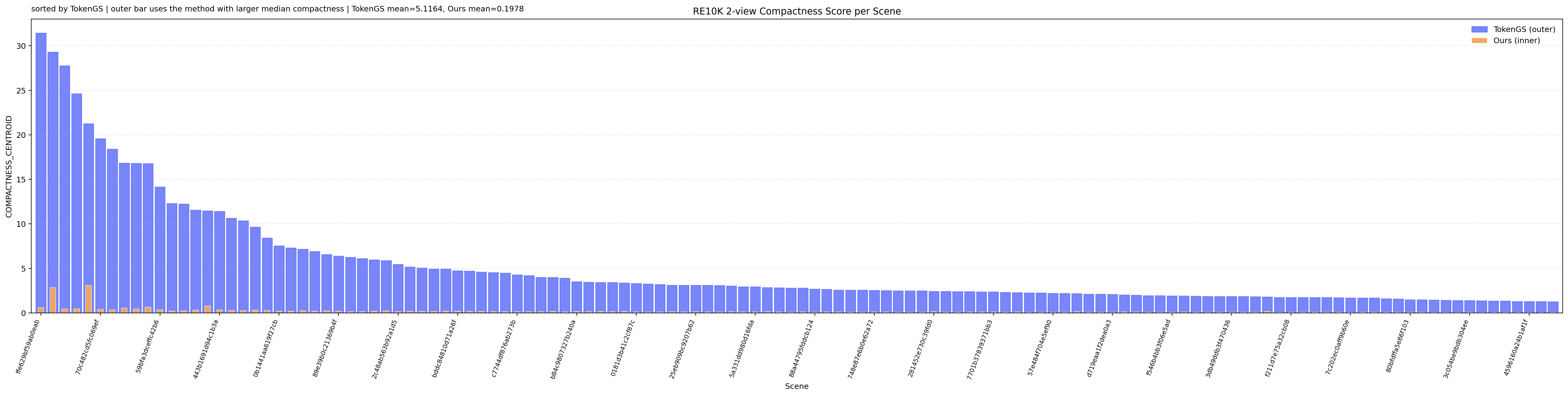}
    \caption{
        \textbf{ Per-scene dispersion score comparison between TokenGS and our method on RE10K.}
        Each overlaid pair of bars corresponds to the same scene. Our method consistently yields substantially lower
         dispersion scores than TokenGS, indicating that the Gaussian groups associated with each anchor are spatially more compact and coherent. 
        }
    \label{fig:re10k_compactness_score}
\end{figure*}

\subsection{More Qualitative Results}
We provide additional qualitative comparisons on DL3DV in \Cref{fig:nvs_and_depth_more_results}. 
LocusGS produces sharper renderings and more coherent depth structures, especially in texture-rich 
regions and cluttered scenes. 

\subsection{Inference Latency and Model Footprint}
We additionally report the model footprint and pure forward latency of LocusGS and TokenGS on the DL3DV 4-view evaluation setup on an NVIDIA A100 40GB GPU. Here, pure forward time measures the encoder-decoder pass from the input views to Gaussian prediction, without including
rendering since both models predict the same number of Gaussians (262,144). Compared with TokenGS, LocusGS increases the parameter count from 222.0M to 241.5M and the model storage footprint from 846.9 MiB to 921.3 MiB.
In terms of speed, TokenGS achieves a mean pure forward time of 341.0 ms per sample, whereas LocusGS requires 407.3 ms.

\begin{table}[t]
\centering
\caption{Model footprint and pure forward latency on the DL3DV 4-view evaluation setup. Pure forward time measures the encoder-decoder pass to Gaussian prediction and excludes rendering.}
\label{tab:efficiency_dl3dv_4view}
\footnotesize
\begin{tabular}{lccc}
\toprule
Model & Params (M) & Model Storage (MiB) & Forward Time (ms) \\
\midrule
TokenGS  & 222.0 & 846.93 & $340.99 \pm 93.57$ \\
LocusGS & 241.5 & 921.31 & $407.28 \pm 62.96$ \\
\bottomrule
\end{tabular}
\end{table}

\subsection{Limitation}
LocusGS currently assumes calibrated input views, since its anchor-aware aggregation relies on camera rays derived from known camera parameters. Extending the framework to pose-free inputs or jointly accounting for pose uncertainty would broaden its applicability. In addition, each token is represented by a center and a scalar radius, which provides a compact but isotropic description of local spatial support. More expressive anchor states, such as anisotropic supports or visibility-aware uncertainty, may better capture elongated structures and geometrically complex regions.

}

\newpage
{
\bibliographystyle{unsrt} 
\bibliography{main,refs_my,refs,vedaldi_general,vedaldi_specific}

@String(CVPR= {IEEE Conf. Comput. Vis. Pattern Recog.})

@String(ICCV= {Int. Conf. Comput. Vis.})

@String(ECCV= {Eur. Conf. Comput. Vis.})

@String(TOG= {ACM Trans. Graph.})

@String(CVPR  = {CVPR})

@String(ICCV  = {ICCV})

@String(ECCV  = {ECCV})

@String(TOG   = {ACM TOG})

@article{loshchilov2017decoupled,
  title   = {Decoupled weight decay regularization},
  author  = {Loshchilov, Ilya and Hutter, Frank},
  journal = {arXiv preprint arXiv:1711.05101},
  year    = {2017}
}

@article{kerbl3Dgaussians,
  author  = {Kerbl, Bernhard and Kopanas, Georgios and Leimk{\"u}hler, Thomas and Drettakis, George},
  title   = {3D Gaussian Splatting for Real-Time Radiance Field Rendering},
  journal = {ACM Transactions on Graphics},
  number  = {4},
  volume  = {42},
  month   = {July},
  year    = {2023},
  url     = {https://repo-sam.inria.fr/fungraph/3d-gaussian-splatting/}
}

@article{mildenhall2021nerf,
  title     = {Nerf: Representing scenes as neural radiance fields for view synthesis},
  author    = {Mildenhall, Ben and Srinivasan, Pratul P and Tancik, Matthew and Barron, Jonathan T and Ramamoorthi, Ravi and Ng, Ren},
  journal   = {Communications of the ACM},
  volume    = {65},
  number    = {1},
  pages     = {99--106},
  year      = {2021},
  publisher = {ACM New York, NY, USA}
}

@article{re10k,
  title   = {Stereo magnification: Learning view synthesis using multiplane images},
  author  = {Zhou, Tinghui and Tucker, Richard and Flynn, John and Fyffe, Graham and Snavely, Noah},
  journal = {arXiv preprint arXiv:1805.09817},
  year    = {2018}
}

@inproceedings{ling2024dl3dv,
  title     = {Dl3dv-10k: A large-scale scene dataset for deep learning-based 3d vision},
  author    = {Ling, Lu and Sheng, Yichen and Tu, Zhi and Zhao, Wentian and Xin, Cheng and Wan, Kun and Yu, Lantao and Guo, Qianyu and Yu, Zixun and Lu, Yawen and others},
  booktitle = {Proceedings of the IEEE/CVF Conference on Computer Vision and Pattern Recognition},
  pages     = {22160--22169},
  year      = {2024}
}

@inproceedings{
  liu2022dabdetr,
  title={{DAB}-{DETR}: Dynamic Anchor Boxes are Better Queries for {DETR}},
  author={Shilong Liu and Feng Li and Hao Zhang and Xiao Yang and Xianbiao Qi and Hang Su and Jun Zhu and Lei Zhang},
  booktitle={International Conference on Learning Representations},
  year={2022},
  url={https://openreview.net/forum?id=oMI9PjOb9Jl}
}

@article{zhang2026anchorsplat,
  title={AnchorSplat: Feed-Forward 3D Gaussian Splatting with 3D Geometric Priors},
  author={Zhang, Xiaoxue and Zheng, Xiaoxu and Yin, Yixuan and Zhao, Tiao and Tang, Kaihua and Mi, Michael Bi and Xu, Zhan and Chen, Dave Zhenyu},
  journal={arXiv preprint arXiv:2604.07053},
  year={2026}
}

@inproceedings{scaffoldgs,
  title={Scaffold-gs: Structured 3d gaussians for view-adaptive rendering},
  author={Lu, Tao and Yu, Mulin and Xu, Linning and Xiangli, Yuanbo and Wang, Limin and Lin, Dahua and Dai, Bo},
  booktitle={Proceedings of the IEEE/CVF Conference on Computer Vision and Pattern Recognition},
  pages={20654--20664},
  year={2024}
}

@article{an2025c3g,
  title={C3G: Learning Compact 3D Representations with 2K Gaussians},
  author={An, Honggyu and Jung, Jaewoo and Kim, Mungyeom and Hong, Sunghwan and Kim, Chaehyun and Fukuda, Kazumi and Jeon, Minkyeong and Han, Jisang and Narihira, Takuya and Ko, Hyuna and others},
  journal={arXiv preprint arXiv:2512.04021},
  year={2025}
}

@InProceedings{tokensplat,
    author    = {Li, Yihui and Lv, Chengxin and Tang, Zichen and Yang, Hongyu and Huang, Di},
    title     = {TokenSplat: Token-aligned 3D Gaussian Splatting for Feed-forward Pose-free Reconstruction},
    booktitle = {Proceedings of the IEEE/CVF Conference on Computer Vision and Pattern Recognition (CVPR)},
    month     = {June},
    year      = {2026},
    pages     = {40886-40895}
}

@inproceedings{lin2025longsplat,
  title={LongSplat: Robust Unposed 3D Gaussian Splatting for Casual Long Videos},
  author={Chin-Yang Lin and Cheng Sun and Fu-En Yang and Min-Hung Chen and Yen-Yu Lin and Yu-Lun Liu},
  booktitle={ICCV},
  year={2025}
}

@article{wang2025zpressor,
  title={ZPressor: Bottleneck-Aware Compression for Scalable Feed-Forward 3DGS},
  author={Wang, Weijie and Chen, Donny Y and Zhang, Zeyu and Shi, Duochao and Liu, Akide and Zhuang, Bohan},
  journal={arXiv preprint arXiv:2505.23734},
  year={2025}
}

@article{wang2024freesplat,
  title={FreeSplat: Generalizable 3D Gaussian Splatting Towards Free-View Synthesis of Indoor Scenes},
  author={Wang, Yunsong and Huang, Tianxin and Chen, Hanlin and Lee, Gim Hee},
  journal={arXiv preprint arXiv:2405.17958},
  year={2024}
}

@inproceedings{xu2024depthsplat,
      title   = {DepthSplat: Connecting Gaussian Splatting and Depth},
      author  = {Xu, Haofei and Peng, Songyou and Wang, Fangjinhua and Blum, Hermann and Barath, Daniel and Geiger, Andreas and Pollefeys, Marc},
      booktitle={CVPR},
      year={2025}
    }

@article{globalsplat,
  title   = {GlobalSplat: Efficient Feed-Forward 3D Gaussian Splatting via Global Scene Tokens},
  author  = {Itkin, Roni and Issachar, Noam and Keypur, Yehonatan and Chen, Xingyu and Chen, Anpei and Benaim, Sagie},
  journal = {arXiv preprint arXiv:2604.15284},
  year    = {2026}
}

@article{tokengs2026,
  title={TokenGS: Decoupling 3D Gaussian Prediction from Pixels with Learnable Tokens},
  author={Jiawei Ren and Michal Tyszkiewicz and Jiahui Huang and Zan Gojcic},
  journal={Proceedings of the IEEE/CVF Conference on Computer Vision and Pattern Recognition},
  year={2026}
}

@article{gslrm2024,
    author={Zhang, Kai and Bi, Sai and Tan, Hao and Xiangli, Yuanbo and Zhao, Nanxuan 
      and Sunkavalli, Kalyan and Xu, Zexiang},
    title     = {GS-LRM: Large Reconstruction Model for 3D Gaussian Splatting},
    journal   = {European Conference on Computer Vision},
    year      = {2024},
}

@article{jiang2025anysplat,
  title={Anysplat: Feed-forward 3d gaussian splatting from unconstrained views},
  author={Jiang, Lihan and Mao, Yucheng and Xu, Linning and Lu, Tao and Ren, Kerui and Jin, Yichen and Xu, Xudong and Yu, Mulin and Pang, Jiangmiao and Zhao, Feng and others},
  journal={ACM Transactions on Graphics (TOG)},
  volume={44},
  number={6},
  pages={1--16},
  year={2025},
  publisher={ACM New York, NY, USA}
}

@inproceedings{ziwen2025llrm,
  title={Long-LRM: Long-sequence Large Reconstruction Model for Wide-coverage Gaussian Splats},
  author={Ziwen, Chen and Tan, Hao and Zhang, Kai and Bi, Sai and Luan, Fujun and Hong, Yicong and Fuxin, Li and Xu, Zexiang},
  booktitle={Proceedings of the IEEE/CVF International Conference on Computer Vision},
  year={2025}
}

@article{ye2024noposplat,
      title   = {No Pose, No Problem: Surprisingly Simple 3D Gaussian Splats from Sparse Unposed Images},
      author  = {Ye, Botao and Liu, Sifei and Xu, Haofei and Xueting, Li and Pollefeys, Marc and Yang, Ming-Hsuan and Songyou, Peng},
      journal = {arXiv preprint arXiv:2410.24207},
      year    = {2024}
    }

@inproceedings{kang2025selfsplat,
  title={SelfSplat: Pose-free and 3D prior-free generalizable 3D Gaussian splatting},
  author={Kang, Gyeongjin and Yoo, Jisang and Park, Jihyeon and Nam, Seungtae and Im, Hyeonsoo and Shin, Sangheon and Kim, Sangpil and Park, Eunbyung},
  booktitle={Proceedings of the Computer Vision and Pattern Recognition Conference},
  pages={22012--22022},
  year={2025}
}

@STRING{arxiv   = {arXiv} }

@STRING{cvpr    = {Proc. {CVPR}} }

@STRING{eccv    = {Proc. {ECCV}} }

@STRING{iccv    = {Proc. {ICCV}} }

@STRING{tog     = {ACM Trans. on Graphics (TOG)} }

@Article{chen24mvsplat:,
  author     = {Yuedong Chen and Haofei Xu and Chuanxia Zheng and Bohan
               Zhuang and Marc Pollefeys and Andreas Geiger and Tat-Jen Cham
               and Jianfei Cai},
  journal    = arxiv,
  title      = {{MVSplat:} Efficient 3D Gaussian Splatting from Sparse
               Multi-View Images},
  volume     = {2403.14627},
  year       = {2024}
}

@InProceedings{charatan23pixelsplat:,
  author     = {David Charatan and Sizhe Li and Andrea Tagliasacchi and
               Vincent Sitzmann},
  booktitle  = cvpr,
  title      = {{pixelSplat}: {3D Gaussian} Splats from Image Pairs for
               Scalable Generalizable {3D} Reconstruction},
  year       = {2024}
}

@InProceedings{schonberger16structure-from-motion,
  author     = {Sch\"{o}nberger, Johannes Lutz and Frahm, Jan-Michael},
  booktitle  = cvpr,
  title      = {Structure-from-Motion Revisited},
  year       = {2016}
}

@InProceedings{schonberger16pixelwise,
  author     = {Sch\"{o}nberger, Johannes Lutz and Zheng, Enliang and
               Pollefeys, Marc and Frahm, Jan-Michael},
  booktitle  = eccv,
  title      = {Pixelwise View Selection for Unstructured Multi-View Stereo},
  year       = {2016}
}

@STRING{arxiv   = {arXiv preprint} }

@STRING{cvpr    = {Proceedings of the {IEEE} Conference on Computer Vision
                  and Pattern Recognition ({CVPR})} }

@STRING{eccv    = {Proceedings of the European Conference on Computer Vision
                  ({ECCV})} }

@STRING{iccv    = {Proceedings of the International Conference on Computer
                  Vision ({ICCV})} }
}

\end{document}